\documentclass[11pt]{article}

\usepackage[preprint]{acl}

\usepackage{times}
\usepackage{latexsym}

\usepackage[T1]{fontenc}

\usepackage[utf8]{inputenc}

\usepackage{booktabs}
\usepackage{multirow}
\usepackage{amssymb}
\definecolor{codebg}{RGB}{250,250,250}        
\definecolor{codekw}{RGB}{33, 74, 135}        
\definecolor{codecomment}{RGB}{63, 127, 95}    
\definecolor{codestring}{RGB}{163, 21, 21}     
\definecolor{codeid}{RGB}{0,0,0}
\usepackage{xcolor}    
\usepackage{listings}  
\usepackage{tcolorbox} 
\tcbuselibrary{listings}
\lstdefinestyle{mypython}{
	language=Python,
	backgroundcolor=\color{codebg},
	basicstyle=\ttfamily\footnotesize,
	keywordstyle=\color{codekw}\bfseries,
	commentstyle=\color{codecomment}\itshape,
	stringstyle=\color{codestring},
	identifierstyle=\color{codeid},
	showstringspaces=false,
	breaklines=true,
	tabsize=4,
	numbers=none,
	numberstyle=\tiny\color{gray},
	stepnumber=1,
	numbersep=8pt,
	frame=single,
	rulecolor=\color{gray},
	frameround=tttt
}
\newtcblisting{codebox}[2][]{
	listing only,
	listing options={style=mypython},
	enhanced,
	colback=mycream,
	colframe=mybrown,
	fonttitle=\bfseries,
	colbacktitle=mybrown,
	coltitle=white,
	title=#2,
	boxed title style={arc=0mm, outer arc=1mm,width=\linewidth},
	arc=0mm,
	left=0mm,
	#1
}
\usepackage{microtype}

\usepackage{inconsolata}

\usepackage{graphicx}

\title{AutoSG: LLM-Driven Solver Generation Solely from Task Prompts for Expensive Optimization}
\author{
	\textbf{Haoran Gu}\textsuperscript{$\clubsuit$} \;\;\;  
	\textbf{Handing Wang}\textsuperscript{$\clubsuit$} \;\;\;
	\textbf{Yi Mei}\textsuperscript{$\diamondsuit$} \;\;\; 
	\textbf{Mengjie Zhang}\textsuperscript{$\diamondsuit$}\;\;\; \\
	\textsuperscript{$\clubsuit$}Xidian University \; 
	\textsuperscript{$\diamondsuit$}Victoria University of Wellington \\
	\texttt{xdu\_guhaoran@163.com}, \texttt{hdwang@xidian.edu.cn}, \texttt{\{yi.mei,mengjie.zhang\}@ecs.vuw.ac.nz}
}

\usepackage{algorithm}
\usepackage{algorithmic}
\usepackage{latexsym}
\usepackage{enumitem}
\usepackage{diagbox}
\usepackage{colortbl}
\usepackage{subcaption}              
\usepackage[utf8]{inputenc}
\usepackage{microtype}
\usepackage{makecell}
\usepackage{multirow}                
\usepackage{booktabs}
\usepackage{tcolorbox}
\tcbuselibrary{breakable} 
\usepackage{caption}      
\usepackage{svg}
\usepackage[subrefformat=parens,labelformat=parens]{subcaption}  
\usepackage{inconsolata}
\usepackage{fontawesome5}
\usepackage{booktabs}
\usepackage{tabularx}
\usepackage{bm} 
\usepackage{amsmath}
\usepackage{tabularx}
\usepackage{adjustbox}
\usepackage{forest}
\usepackage{lipsum}
\usepackage{amsfonts}
\usepackage{multirow,booktabs,color,soul,threeparttable}
\PassOptionsToPackage{table,xcdraw}{xcolor}
\usepackage{xcolor}
\usepackage{colortbl}
\tcbuselibrary{listings, breakable} 
\usepackage[table]{xcolor}
\definecolor{lightpink}{RGB}{255, 230, 235}
\definecolor{lightblue}{RGB}{230, 240, 255}
\tcbset{
	userstyle/.style={
		enhanced,
		breakable,
		colback=white,
		colframe=black,
		colbacktitle=gray!20,
		coltitle=black,
		rounded corners,
		sharp corners=north,
		boxrule=0.5pt,
		drop shadow=black!50!white,
		attach boxed title to top left={
			xshift=-2mm,
			yshift=-2mm
		},
		boxed title style={
			rounded corners,
			size=small,
			colback=gray!20
		}
	},
	replystyleg/.style={
		enhanced,
		breakable,
		colback=green!15,
		colframe=black,
		colbacktitle=green!30,
		coltitle=black,
		boxrule=0.5pt,
		drop shadow=black!50!white,
		rounded corners,
		sharp corners=north,
		attach boxed title to top right={
			xshift=-2mm,
			yshift=-2mm
		},
		boxed title style={
			rounded corners,
			size=small,
			colback=green!40
		}
	},
	replystyler/.style={
		enhanced,
		breakable,
		colback=red!15,
		colframe=black,
		colbacktitle=red!40,
		coltitle=black,
		boxrule=0.5pt,
		drop shadow=black!50!white,
		rounded corners,
		sharp corners=north,
		attach boxed title to top right={
			xshift=-2mm,
			yshift=-2mm
		},
		boxed title style={
			rounded corners,
			size=small,
			colback=red!40
		}
	}
}

\newtcolorbox{userquery}[1][]{
	userstyle,
	title=Prompt,
	#1
}

\newtcolorbox{llmreply-g}[1][]{
	replystyleg,
	title=Response,
	#1
}

\newtcolorbox{llmreply-r}[1][]{
	replystyler,
	title=Response,
	#1
}

\begin{document}
\maketitle
\begin{abstract}
	Expensive optimization tasks are ubiquitous in real-world applications, demanding highly specialized solvers. While LLM-driven automated solver generation shows promise, current paradigms face three critical issues when tackling expensive optimization: factual hallucinations due to deficient domain knowledge, the frequent dismantling of previously established locally optimal structures during refinement, and the prohibitive evaluation costs alongside restricted generalization caused by executing on training instances. To address these issues, we introduce AutoSG, a fully automated workflow directly translating natural language prompts into executable customized solvers. AutoSG features three core innovations: a retrieval-augmented solver generation module strictly grounding code in verified literature; a one-step self-refinement operator introducing task-specific improvements while preserving critical structural components; and an instance-free Elo-based LLM-as-a-Judge evaluation mechanism rapidly establishing global rankings. Extensive evaluations across diverse expensive optimization tasks confirm AutoSG significantly outperforms human-designed state-of-the-art frameworks and existing LLM-generated solvers.
\end{abstract}

\section{Introduction} \label{s1}
Expensive optimization problems \cite{8456559,9971764} are prevalent in domains like neural network hyperparameter tuning \cite{liu2024large}, antenna design \cite{11386963}, and airfoil shape optimization \cite{8357456}. Solving these tasks involves finding optimal parameter configurations for peak performance. However, evaluating any configuration typically requires high-fidelity simulations or destructive physical prototyping \cite{2022cost}. Given their prohibitive time or financial costs, traditional solvers relying on massive trial-and-error iterations are entirely infeasible. Consequently, mitigating this bottleneck necessitates highly customized, data-efficient solvers based on Bayesian optimization \cite{wang2023recent} or surrogate-assisted evolutionary frameworks \cite{jin2011surrogate} to navigate complex search spaces within a strictly limited evaluation budget.

Conventionally, the design of such solvers relies heavily on human expertise, which is inherently time-consuming and labor-intensive. Recently, large language models (LLMs) \cite{gu2026overlooked,helpfulass} have demonstrated immense potential in code generation, driving a paradigm shift in automated solver design from zero-shot generation to evolutionary iterative frameworks \cite{liu2026systematic}.
However, applying these current LLM-driven paradigms to design complex expensive optimization solvers presents three challenges. First, regarding the initial generation phase, LLMs lack sufficient domain-specific pre-trained knowledge. Therefore, they are prone to severe factual hallucinations \cite{zhang2025llm}, making it fundamentally difficult to generate high-quality and structurally sound solvers. Second, during the refinement phase, existing operators tend to catastrophically destroy the previously established locally optimal structures, making the quality of offspring solvers exceptionally difficult to guarantee. Third, concerning the evaluation phase, relying on actual executions on training instances introduces two severe limitations. Specifically, the prohibitive cost of evaluation on these expensive instances renders the trial-and-error assessments demanded by evolutionary iterations or even basic validation fundamentally infeasible \cite{yin2026landscape}. For example, assessing a single solver for antenna design under a budget of merely 300 function evaluations (FEs) can consume 75 hours of simulation time \cite{11386963}. Additionally, the performance feedback derived from specific training instances suffers from limited generalization when applied to unseen instances.

To address these challenges, we propose AutoSG, a novel LLM-driven automated solver generation workflow tailored for expensive optimization problems. By seamlessly integrating a retrieval-augmented generation (RAG)-based solver generation engine, a one-step solver self-refinement operator, and an Elo-based LLM-as-a-Judge evaluation mechanism, AutoSG achieves a groundbreaking leap: directly translating natural language task prompts into executable and highly customized optimization solvers. The main contributions of this work are summarized as follows:

\begin{itemize}[leftmargin=*]
	\item We introduce AutoSG, a novel LLM-driven workflow for expensive optimization. Empowered by a fully generalized template, AutoSG demonstrates exceptional versatility, directly translating natural language task prompts into executable solvers dynamically tailored to accommodate varying search space dimensions and evaluation budget constraints.
	
	\item We propose a RAG-based solver generation engine that autonomously parses user queries to retrieve specialized domain literature from external academic databases. By adopting a two-stage LLM-based generation process strictly grounded in verified literature, this engine effectively eliminates factual hallucinations and reliably produces state-of-the-art (SOTA) level solvers.
	
	\item We design a structurally safe one-step self-refinement operator to further improve the generated solvers. This operator introduces task-specific enhancements with minimal API invocation overhead while meticulously preserving the previously established locally optimal structures.
	
	\item We integrate an efficient Elo-based LLM-as-a-Judge evaluation mechanism to rapidly establish reliable global rankings and identify the final optimal solver. This novel instance-free solver evaluation paradigm entirely bypasses any actual executions on training instances.
\end{itemize}
\section{Related Work}
\subsection{LLM-Driven Solver Generation}
\begin{figure*}[t]
	\centering
	\includegraphics[width=\textwidth]{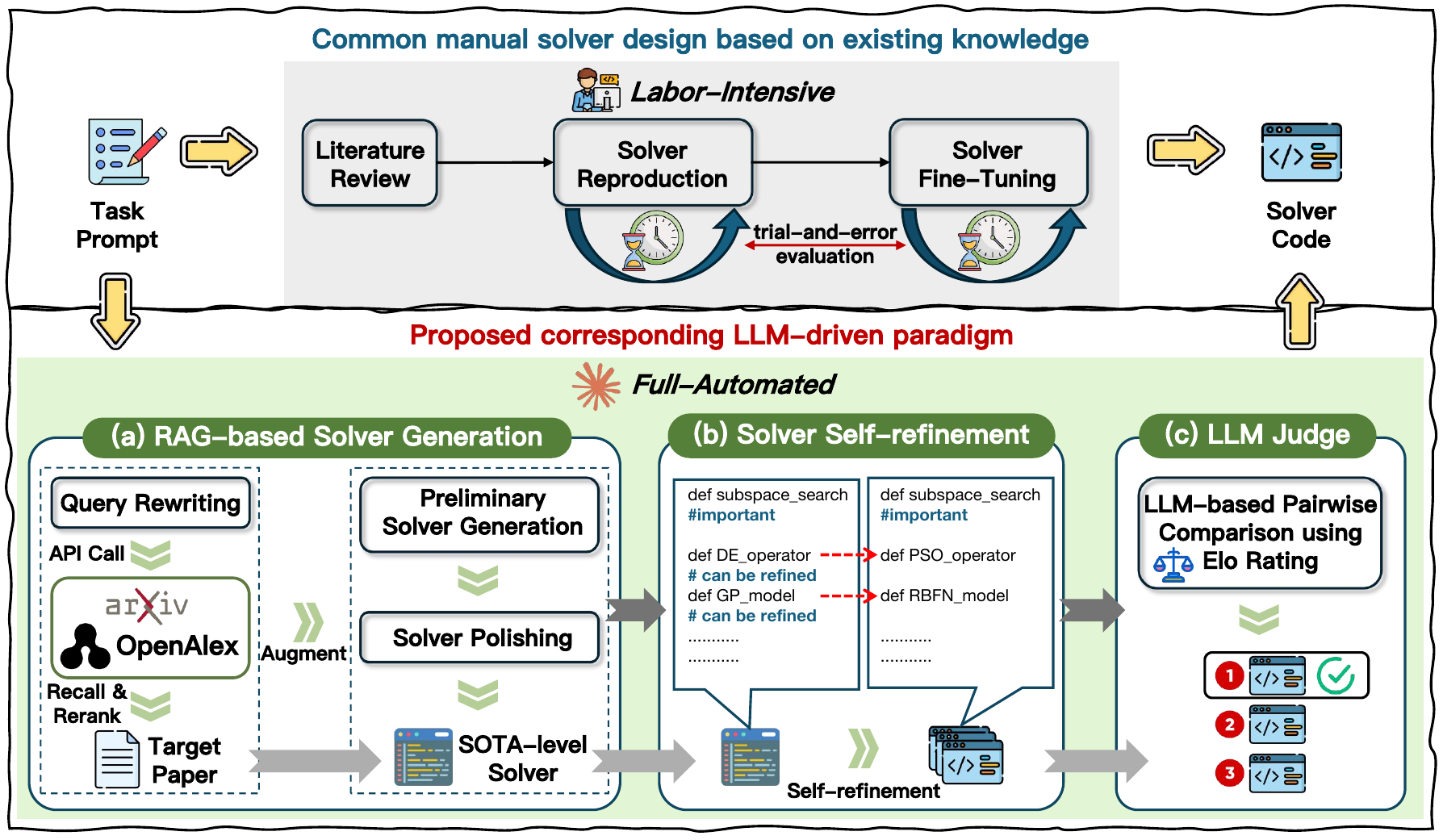}
	\caption{The flowchart of AutoSG, in contrast with the corresponding manual solver design based on existing knowledge.}
	\label{fig1}
\end{figure*}

For combinatorial optimization and mathematical reasoning, several pioneering frameworks demonstrate the potential of LLMs as heuristic generators. FunSearch \cite{romera2024mathematical} couples an LLM with an automated evaluation loop to discover novel mathematical constructions. Similarly, EoH \cite{fei2024eoh} demonstrates that LLMs can seamlessly replace traditional variation operators by intelligently mutating and recombining executable heuristic programs. Building upon this, ReEvo \cite{ye2024reevo} introduces a hierarchical reflective mechanism, extracting explicit design principles from past historical performance to guide a more directed search process. Finally, expanding beyond isolated algorithmic operators, LLaMEA \cite{van2025llamea} proves the feasibility of evolving complete, standalone metaheuristic solvers through a generate-evaluate-improve cycle.

Recently, these generative paradigms have been extended to expensive optimization. For instance, EvolCAF \cite{yao2024evolve} integrates LLMs with evolutionary computation to automatically design novel, cost-aware acquisition functions, iteratively evolving their mathematical descriptions and code blocks. 
While this method enhances specific components, LLaMEA-BO \cite{li2025llamea} automatically generates and iteratively refines complete Bayesian optimization Python solvers (including initial sampling, surrogate, and acquisition function) via continuous benchmark evaluations.

Despite these advancements, there are three critical limitations when tackling expensive optimization. First, generating solvers from scratch exposes the LLM's lack of specialized empirical knowledge, fundamentally constraining the structural soundness and quality of the generated solvers. Second, existing refinement operators frequently dismantle the previously established locally optimal structures, resulting in offspring solvers of deteriorated quality. Third, the common reliance on actual executions on training instances introduces a dual challenge. On one hand, the prohibitive computational cost associated with these evaluations on expensive problems renders intensive trial-and-error iterations completely infeasible. On the other hand, the performance feedback derived from such specific cases can restrict the generalization capability of the solvers across unseen scenarios.
As a result, systematically resolving these exact limitations constitutes the primary goal of our work.

\subsection{LLM-as-a-Judge}
The LLM-as-a-Judge paradigm, initially introduced to bypass expensive human annotations in natural language processing \cite{li2025generation,zheng2023judging}, has recently emerged as a powerful solution for automated code assessment \cite{jiang2025codejudgebench,he2025code,zhao2025codejudge}. Unlike traditional metrics requiring explicit reference implementations, advanced LLMs leverage their deep programming comprehension and chain-of-thought reasoning to directly evaluate algorithm quality. This pure static verification completely avoids the prohibitive computational costs associated with dynamic execution environments. Motivated by this, we adopt this robust paradigm as the foundation for our proposed entirely instance-free evaluation mechanism.
\section{AutoSG Workflow}
\label{sec:method}
To overcome the limitations of existing LLM-driven automated paradigms, we draw an inspiration from the systematic workflow of human domain experts. In practice, when confronted with a specialized optimization task, experts do not invent solvers out of thin air \cite{human, benureau2018re}. Instead, they begin with an extensive literature review to identify the SOTA-level solver tailored to the problem, \textbf{which naturally circumvents the knowledge deficiency issue}. Subsequently, experts manually reproduce this SOTA-level solver into executable code and meticulously fine-tune (\textbf{rather than causing catastrophic structural destruction}) its internal mechanisms to enhance problem-solving capabilities.

Our primary motivation is to fully automate this exact pipeline, utilizing the advanced capabilities of LLMs to seamlessly mimic and replace human intervention across all stages. Driven by this vision, we introduce AutoSG, a fully LLM-driven automated workflow for expensive optimization solver generation. 
As Figure~\ref{fig1} illustrates, the pipeline processes a user-provided task prompt through three core modules: a RAG-based solver generation engine, a one-step self-refinement operator, and an Elo-based LLM-as-a-Judge evaluation mechanism. This systematic execution finally outputs an executable solver perfectly tailored to the target optimization task.
\subsection{Task Prompt}
The primary input to AutoSG is a user-defined task prompt, denoted as $\mathcal{P}_{task}$, which encapsulates the essential characteristics and constraints of the target expensive optimization problem. Formally, we define the task prompt as a tuple:
\begin{equation}
	\mathcal{P}_{task} = \langle \mathcal{T}_{desc}, d, B, \mathcal{X} \rangle
\end{equation}
where the components are defined as follows:
\begin{itemize}[leftmargin=*]
	\item $\mathcal{T}_{desc}$ represents a concise natural language description of the optimization task or objective function characteristics.
	\item $d$ denotes the dimensionality of the problem's decision variables.
	\item $B$ specifies the strictly limited budget of expensive FEs.
	\item $\mathcal{X}$ defines the search space or the boundary constraints for the decision variables.
\end{itemize}

Given this comprehensive prompt, the proposed AutoSG workflow acts as an automated generative mapping. It generates and refines algorithmic logic to output a highly customized, executable solver $\mathcal{S}^*$. This end-to-end transformation is formulated as:
\begin{equation}
	\mathcal{S}^* \leftarrow \text{AutoSG}(\mathcal{P}_{task})
\end{equation}
where $\mathcal{S}^*$ is the final generated solver. Notably, the output solver is already instantiated and optimized for the specific dimension $d$ and evaluation budget $B$, making it immediately deployable for the target black-box evaluation environment.
\subsection{RAG-based Solver Generation}
\label{sec32}
Automating the literature review and solver generation process with LLMs presents two distinct paradigms. The first is an ungrounded approach, prompting the LLM to self-recommend literature relying solely on its internal parametric memory. Formally, by concatenating the problem definition $\mathcal{P}_{task}$ with a meta-query $\mathcal{Q}_{meta}$ (e.g., \textit{``Recommend the most relevant state-of-the-art literature for this optimization task and construct its executable solver code''}), this zero-shot paradigm is expressed as:
\begin{equation}
	\tilde{\mathcal{S}} \leftarrow \text{LLM}(\mathcal{P}_{task} \oplus \mathcal{Q}_{meta})
		\label{eqdllm}
\end{equation}
where $\oplus$ denotes prompt concatenation, and $\tilde{\mathcal{S}}$ represents the naively generated solver. 

However, this paradigm is highly susceptible to factual hallucinations, manifesting in two critical failure modes: (1) \textbf{Source Fabrication}: the model invents non-existent literature or recommends misaligned backgrounds; and (2) \textbf{Implementation Degradation}: lacking granular mathematical details within its parametric memory, the LLM often reproduces structurally flawed or sub-optimal code.
To circumvent these limitations, we adopt the second paradigm: explicitly grounding the LLM with an external knowledge base.

\noindent \textbf{Retrieval phase.} First, the LLM rewrites the task prompt into a diverse set of $K$ domain-specific search queries, formulated as $\mathcal{Q}_{exp} = \{q_1, \dots, q_K\} \leftarrow \text{LLM}(\mathcal{P}_{rewrite} \oplus \mathcal{P}_{task})$, where $\mathcal{P}_{rewrite}$ is the instructional prompt. Second, we leverage $\mathcal{Q}_{exp}$ to simultaneously search the OpenAlex and arXiv academic databases via their APIs. For each query, we recall the top $M_{\text{OA}}$ and $M_{\text{ax}}$ paper records from OpenAlex and arXiv respectively, ranked by relevance, yielding an initial retrieval pool of up to $K \times (M_{\text{OA}} + M_{\text{ax}})$ papers. To ensure source balance and temporal freshness, the OpenAlex and arXiv pools are independently deduplicated and sorted by publication year, retaining the latest $N_{\text{OA}}$ and $N_{\text{ax}}$ candidate papers, respectively. These two sets are subsequently merged to form a unified pool of $N$ candidates. Finally, the textual metadata (e.g., titles, citations, venues, and full abstracts) of the $N$ candidates is fed into the LLM for reranking to identify the single most relevant paper. The full-text PDF file of this Top-1 paper is then parsed and extracted as the final retrieved content, denoted as $\mathcal{D}_{pdf}$:
\begin{equation}
	\mathcal{D}_{pdf} \leftarrow \text{LLM}(\mathcal{P}_{rerank} \oplus \{meta_1, \dots, meta_N\})
\end{equation}
where $\mathcal{P}_{rerank}$ serves as the reranking prompt and $meta_i$ represents the textual metadata of the $i$-th candidate paper. 

\noindent \textbf{Solver generation phase.} Equipped with the retrieved empirical knowledge $\mathcal{D}_{pdf}$, we employ a two-stage generative pipeline to instantiate the unstructured academic text into an executable solver code. In the first stage, the LLM acts as a faithful reproducer to generate a preliminary solver, denoted as $\mathcal{S}_{pre}$. This step is driven by a primary system prompt $\mathcal{P}_{G1}$, a highly extensible predefined structural code template $\mathcal{T}_{code}$ capable of instantiating both Bayesian optimization and surrogate-assisted solvers, and a specialized instruction $\mathcal{P}_{imp}$ that mandates the LLM to identify and explicitly annotate critical solver components within the generated code. Formally:
\begin{equation}
	\mathcal{S}_{pre} \leftarrow \text{LLM}(\mathcal{P}_{G1} \oplus \mathcal{D}_{pdf} \oplus \mathcal{T}_{code} \oplus \mathcal{P}_{imp})
\end{equation}
In the second stage, to eliminate any microscopic implementation discrepancies, we introduce a targeted polishing mechanism. Guided by a secondary prompt $\mathcal{P}_{G2}$, the LLM rigorously cross-references the preliminary code $\mathcal{S}_{pre}$ with the native pseudo-code, mathematical formulas, and hyper-parameters detailed in $\mathcal{D}_{pdf}$. This generates initial and SOTA-level solver $\mathcal{S}_{init}$:
\begin{equation}
	\mathcal{S}_{init} \leftarrow \text{LLM}(\mathcal{P}_{G2} \oplus \mathcal{D}_{pdf} \oplus \mathcal{S}_{pre})
	\label{eqinit}
\end{equation}

\subsection{One-Step Solver Self-Refinement}
\label{sec33}
Since the initial solver $\mathcal{S}_{init}$ is explicitly grounded in SOTA-level literature, it already resides near a local optimum within the algorithmic search space. The conventional multi-step iterative evolution is computationally redundant and risks catastrophic forgetting of the original mathematical structures. To efficiently adapt $\mathcal{S}_{init}$ to the specific target problem $\mathcal{P}_{task}$, we introduce a one-step self-refinement operator. First, to establish an explicit semantic understanding of the underlying solver structure, the LLM reverse-engineers $\mathcal{S}_{init}$ with $\mathcal{P}_{reverse}$ to generate a semantic code description $\mathcal{D}_{init}$. Then, guided by a refinement prompt $\mathcal{P}_{refine}$, the LLM adapts the implementation to $\mathcal{P}_{task}$. Crucially, $\mathcal{P}_{refine}$ adopts a ``learn-and-improve'' approach, mandating the LLM to introduce necessary improvements while strictly minimizing modifications to the critical solver components previously annotated via $\mathcal{P}_{imp}$. After that, this operator directly generates a diverse pool of $C$ candidate solvers:
\begin{equation}
	\begin{split}
		\{\mathcal{S}_{L1}, \dots, \mathcal{S}_{LC}\} \leftarrow \text{LLM} ( & \mathcal{P}_{task} \oplus \mathcal{P}_{refine} \oplus \mathcal{T}_{code} \\ \oplus 
		& \mathcal{S}_{init} \oplus \mathcal{D}_{init} )
	\end{split}
\end{equation}
where $C$ is the total number of refined candidates, which are subsequently forwarded to the evaluation module.

\subsection{Elo-based LLM-as-a-Judge Evaluation}
\label{sec34}
The preceding generation phase yields a diverse pool of candidate solvers, denoted as $\mathcal{S}_{candidate} = \{\mathcal{S}_{init}, \mathcal{S}_{L1}, \dots, \mathcal{S}_{LC}\}$. To achieve an entirely instance-free evaluation paradigm, we adopt the LLM-as-a-Judge framework. Specifically, we employ a dynamically updated Elo rating system to drive multi-round pair-wise comparisons, overcoming the inherent score inflation of point-wise methods. Furthermore, this Elo strategy avoids exhaustive pairings, efficiently establishing global rankings with significantly fewer evaluations \cite{chiang2024chatbot}, while its statistical aggregation inherently mitigates LLM randomness and position biases \cite{boubdir2024elo}.

In all matches, the comparative evaluation to determine the superior solver is driven by the adjudication prompt $\mathcal{P}_{judge}$, which dynamically integrates $\mathcal{P}_{task}$ to firmly anchor the judgment on specific task constraints. 
Each solver $\mathcal{S}_i \in \mathcal{S}_{candidate}$ initializes with a rating $R_i^{(0)}$ and deviation $\text{rd}_i^{(0)}$. For a match between winner $\mathcal{S}_w$ and loser $\mathcal{S}_\ell$, ratings update independently via $\Delta R_w = K_w(1 - E_{w,\ell})$ and $\Delta R_\ell = -K_\ell E_{\ell,w}$, where $E_{w,\ell} = 1 / (1 + 10^{(R_\ell - R_w) / 400})$. To accelerate early convergence, we dynamically scale the multiplier $K_i$ based on the solver's historical match count $m_i$:
\begin{equation}
	K_i = 32 \times \left(1 + \frac{50}{m_i + 10}\right)
\end{equation}
After each match, $\text{rd}$ decays via $\max(30, 0.95 \cdot \text{rd})$, forming an approximate 95\% confidence interval (CI) $[R_i \pm 1.96 \cdot \text{rd}_i]$.

To maximize the informativeness of each comparison and accelerate Elo convergence, we execute a two-phase smart-pairing tournament. \textbf{Phase 1} targets a minimum baseline of $T_{init}$ matches per solver, prioritizing pairs with the fewest combined matches: $\text{prio}_{ij} = -(m_i + m_j) + \epsilon$ ($\epsilon \sim U(0, 0.1)$). \textbf{Phase 2} extends the individual target to $T_{total}$ matches, shifting the priority to a composite heuristic balancing rating proximity and evaluation scarcity:
\begin{equation}
	\text{prio}_{ij} = \frac{1000}{|R_i - R_j| + 1} + \lambda \cdot \left(T_{total} - \frac{m_i + m_j}{2}\right)
\end{equation}
where $\lambda$ is a balancing coefficient. Crucially, redundant matchups are aggressively pruned: if their absolute rating difference $|R_i - R_j| > \Delta_{\max}$ and their 95\% CIs are disjoint, the match is skipped.
Ultimately, once all candidates have accumulated a minimum of $T_{total}$ matches, the tournament concludes, and the candidate securing the highest final Elo rating is explicitly selected as the optimal solver $\mathcal{S}^*$.

\section{Experiments}
\label{sec:experiments}
\begin{table*}[htbp]
	\centering
	\resizebox{\textwidth}{!}{
		\setlength{\tabcolsep}{2pt} 
		\begin{tabular}{lccccccc}
			\toprule
			\multirow{2}{*}{\Large \textbf{Pro}} & \multicolumn{4}{c}{\cellcolor{gray!8}\Large \textbf{Baselines}} & \multicolumn{3}{c}{\cellcolor{cyan!10}\textbf{\Large Proposed AutoSG}} \\
			\cmidrule(lr){2-5} \cmidrule(lr){6-8}
			& \cellcolor{gray!8}CMA-ES & \cellcolor{gray!8}TuRBO & \cellcolor{gray!8}HEBO & \cellcolor{gray!8}TREvol & 
			\cellcolor{cyan!10}$\mathcal{S}^*_{\text{BBOB}, 1}$ & \cellcolor{cyan!10}$\mathcal{S}^*_{\text{BBOB}, 2}$ & \cellcolor{cyan!10}$\mathcal{S}^*_{\text{BBOB}, 3}$ \\
			\midrule
			F1 & \cellcolor{gray!8}-1.25e+02(7.9e+00) & \cellcolor{gray!8}-1.50e+02(1.1e+00) & \cellcolor{gray!8}-1.51e+02(1.9e-01) & \cellcolor{gray!8}\textbf{-1.52e+02(2.5e-03)} & \cellcolor{cyan!10}-1.52e+02(1.3e-02) & \cellcolor{cyan!10}-1.52e+02(6.1e-02) & \cellcolor{cyan!10}-1.52e+02(6.3e-02) \\
			F2 & \cellcolor{gray!8}5.06e+05(1.1e+05) & \cellcolor{gray!8}6.00e+04(1.8e+04) & \cellcolor{gray!8}2.14e+03(4.9e+02) & \cellcolor{gray!8}4.07e+04(1.1e+04) & \cellcolor{cyan!10}\textbf{1.71e+03(4.6e+02)} & \cellcolor{cyan!10}1.39e+04(3.3e+03) & \cellcolor{cyan!10}1.58e+04(5.3e+03) \\
			F3 & \cellcolor{gray!8}3.27e+02(3.7e+01) & \cellcolor{gray!8}2.28e+02(5.2e+01) & \cellcolor{gray!8}1.35e+02(1.3e+01) & \cellcolor{gray!8}4.21e+02(4.1e+01) & \cellcolor{cyan!10}\textbf{1.10e+02(2.0e+01)} & \cellcolor{cyan!10}1.22e+02(1.8e+01) & \cellcolor{cyan!10}1.39e+02(2.0e+01) \\
			F4 & \cellcolor{gray!8}5.02e+02(7.8e+01) & \cellcolor{gray!8}3.59e+02(4.7e+01) & \cellcolor{gray!8}2.05e+02(2.0e+01) & \cellcolor{gray!8}4.90e+02(4.5e+01) & \cellcolor{cyan!10}\textbf{1.85e+02(2.1e+01)} & \cellcolor{cyan!10}2.32e+02(3.7e+01) & \cellcolor{cyan!10}2.46e+02(3.8e+01) \\
			F5 & \cellcolor{gray!8}1.05e+03(1.5e+01) & \cellcolor{gray!8}9.72e+02(4.4e+00) & \cellcolor{gray!8}\textbf{9.44e+02(3.9e-01)} & \cellcolor{gray!8}1.02e+03(1.5e+01) & \cellcolor{cyan!10}9.49e+02(1.8e+00) & \cellcolor{cyan!10}9.52e+02(2.4e+00) & \cellcolor{cyan!10}9.53e+02(2.0e+00) \\
			F6 & \cellcolor{gray!8}3.69e+04(1.4e+04) & \cellcolor{gray!8}2.68e+02(3.3e+01) & \cellcolor{gray!8}2.83e+02(3.5e+01) & \cellcolor{gray!8}2.58e+02(2.9e+01) & \cellcolor{cyan!10}\textbf{1.99e+02(1.6e+01)} & \cellcolor{cyan!10}2.00e+02(1.4e+01) & \cellcolor{cyan!10}2.08e+02(1.8e+01) \\
			F7 & \cellcolor{gray!8}2.19e+02(4.7e+01) & \cellcolor{gray!8}7.06e+01(1.2e+01) & \cellcolor{gray!8}4.98e+01(1.1e+01) & \cellcolor{gray!8}1.87e+02(5.1e+01) & \cellcolor{cyan!10}\textbf{4.43e+01(7.1e+00)} & \cellcolor{cyan!10}4.81e+01(8.4e+00) & \cellcolor{cyan!10}5.47e+01(6.5e+00) \\
			F8 & \cellcolor{gray!8}1.16e+04(3.3e+03) & \cellcolor{gray!8}9.85e+02(3.5e+02) & \cellcolor{gray!8}\textbf{1.57e+01(3.3e+01)} & \cellcolor{gray!8}5.29e+01(2.5e+01) & \cellcolor{cyan!10}3.30e+01(2.1e+01) & \cellcolor{cyan!10}2.05e+02(1.0e+02) & \cellcolor{cyan!10}1.79e+02(5.1e+01) \\
			F9 & \cellcolor{gray!8}5.02e+03(1.3e+03) & \cellcolor{gray!8}9.56e+02(7.1e+02) & \cellcolor{gray!8}5.43e+01(1.3e+02) & \cellcolor{gray!8}7.18e+01(5.7e+01) & \cellcolor{cyan!10}\textbf{-2.21e+01(2.4e+01)} & \cellcolor{cyan!10}1.04e+02(6.5e+01) & \cellcolor{cyan!10}1.49e+02(6.2e+01) \\
			F10 & \cellcolor{gray!8}6.00e+05(1.1e+05) & \cellcolor{gray!8}1.97e+05(4.8e+04) & \cellcolor{gray!8}1.22e+05(5.9e+04) & \cellcolor{gray!8}\textbf{4.39e+04(1.5e+04)} & \cellcolor{cyan!10}4.82e+04(1.6e+04) & \cellcolor{cyan!10}5.91e+04(1.9e+04) & \cellcolor{cyan!10}7.53e+04(2.4e+04) \\
			F11 & \cellcolor{gray!8}1.30e+03(7.6e+01) & \cellcolor{gray!8}1.23e+03(2.8e+01) & \cellcolor{gray!8}1.25e+03(4.9e+01) & \cellcolor{gray!8}\textbf{1.08e+03(1.7e+01)} & \cellcolor{cyan!10}1.15e+03(3.4e+01) & \cellcolor{cyan!10}1.14e+03(2.7e+01) & \cellcolor{cyan!10}1.13e+03(3.5e+01) \\
			F12 & \cellcolor{gray!8}4.06e+07(1.1e+07) & \cellcolor{gray!8}3.55e+07(1.8e+07) & \cellcolor{gray!8}2.16e+05(6.4e+04) & \cellcolor{gray!8}\textbf{1.48e+04(3.7e+03)} & \cellcolor{cyan!10}1.41e+05(7.6e+04) & \cellcolor{cyan!10}4.79e+06(1.5e+06) & \cellcolor{cyan!10}3.87e+06(1.9e+06) \\
			F13 & \cellcolor{gray!8}1.94e+03(1.1e+02) & \cellcolor{gray!8}1.18e+03(4.7e+01) & \cellcolor{gray!8}1.03e+03(2.1e+01) & \cellcolor{gray!8}\textbf{9.81e+02(1.6e+01)} & \cellcolor{cyan!10}9.89e+02(1.2e+01) & \cellcolor{cyan!10}1.02e+03(1.3e+01) & \cellcolor{cyan!10}1.05e+03(1.3e+01) \\
			F14 & \cellcolor{gray!8}5.30e+01(2.9e+00) & \cellcolor{gray!8}4.47e+01(1.2e+00) & \cellcolor{gray!8}4.04e+01(1.5e-01) & \cellcolor{gray!8}\textbf{4.00e+01(7.8e-03)} & \cellcolor{cyan!10}4.01e+01(9.4e-02) & \cellcolor{cyan!10}4.12e+01(5.4e-01) & \cellcolor{cyan!10}4.11e+01(4.7e-01) \\
			F15 & \cellcolor{gray!8}3.52e+02(3.4e+01) & \cellcolor{gray!8}2.80e+02(3.7e+01) & \cellcolor{gray!8}2.07e+02(1.2e+01) & \cellcolor{gray!8}4.52e+02(6.6e+01) & \cellcolor{cyan!10}1.52e+02(2.5e+01) & \cellcolor{cyan!10}\textbf{1.48e+02(2.1e+01)} & \cellcolor{cyan!10}1.69e+02(1.7e+01) \\
			F16 & \cellcolor{gray!8}-4.83e+02(2.4e+00) & \cellcolor{gray!8}-4.99e+02(3.9e+00) & \cellcolor{gray!8}-5.05e+02(2.3e+00) & \cellcolor{gray!8}-4.99e+02(5.2e+00) & \cellcolor{cyan!10}\textbf{-5.09e+02(2.8e+00)} & \cellcolor{cyan!10}-5.05e+02(4.7e+00) & \cellcolor{cyan!10}-5.04e+02(2.8e+00) \\
			F17 & \cellcolor{gray!8}4.57e+01(1.2e+00) & \cellcolor{gray!8}4.43e+01(1.5e+00) & \cellcolor{gray!8}4.12e+01(9.7e-01) & \cellcolor{gray!8}4.46e+01(7.9e-01) & \cellcolor{cyan!10}\textbf{3.97e+01(8.7e-01)} & \cellcolor{cyan!10}4.06e+01(9.0e-01) & \cellcolor{cyan!10}4.09e+01(8.1e-01) \\
			F18 & \cellcolor{gray!8}6.62e+01(2.9e+00) & \cellcolor{gray!8}6.39e+01(4.7e+00) & \cellcolor{gray!8}5.08e+01(4.2e+00) & \cellcolor{gray!8}6.32e+01(2.0e+00) & \cellcolor{cyan!10}\textbf{4.58e+01(2.3e+00)} & \cellcolor{cyan!10}4.71e+01(1.7e+00) & \cellcolor{cyan!10}5.14e+01(4.2e+00) \\
			F19 & \cellcolor{gray!8}-4.93e+01(5.0e-01) & \cellcolor{gray!8}-5.01e+01(9.5e-01) & \cellcolor{gray!8}-5.11e+01(2.2e-01) & \cellcolor{gray!8}-5.15e+01(4.5e-01) & \cellcolor{cyan!10}\textbf{-5.22e+01(4.1e-01)} & \cellcolor{cyan!10}-5.21e+01(3.8e-01) & \cellcolor{cyan!10}-5.18e+01(5.7e-01) \\
			F20 & \cellcolor{gray!8}4.55e+03(2.4e+03) & \cellcolor{gray!8}-1.42e+02(2.8e-01) & \cellcolor{gray!8}-1.42e+02(1.3e-01) & \cellcolor{gray!8}-1.42e+02(2.4e-01) & \cellcolor{cyan!10}-1.42e+02(2.0e-01) & \cellcolor{cyan!10}-1.42e+02(1.8e-01) & \cellcolor{cyan!10}\textbf{-1.43e+02(2.5e-01)} \\
			F21 & \cellcolor{gray!8}1.09e+01(1.1e+01) & \cellcolor{gray!8}-9.29e+00(9.5e+00) & \cellcolor{gray!8}-1.91e+01(1.1e+01) & \cellcolor{gray!8}-2.28e+01(9.7e+00) & \cellcolor{cyan!10}-3.07e+01(7.3e-01) & \cellcolor{cyan!10}\textbf{-3.12e+01(6.1e-01)} & \cellcolor{cyan!10}-3.06e+01(9.1e-01) \\
			F22 & \cellcolor{gray!8}3.53e+02(1.4e+01) & \cellcolor{gray!8}3.10e+02(1.2e+01) & \cellcolor{gray!8}3.08e+02(6.5e+00) & \cellcolor{gray!8}2.99e+02(4.3e-01) & \cellcolor{cyan!10}\textbf{2.99e+02(1.1e-01)} & \cellcolor{cyan!10}2.99e+02(1.5e-01) & \cellcolor{cyan!10}2.99e+02(2.1e-01) \\
			F23 & \cellcolor{gray!8}-2.18e+02(4.6e-01) & \cellcolor{gray!8}-2.18e+02(4.7e-01) & \cellcolor{gray!8}-2.19e+02(3.5e-01) & \cellcolor{gray!8}-2.18e+02(4.0e-01) & \cellcolor{cyan!10}-2.19e+02(4.0e-01) & \cellcolor{cyan!10}-2.20e+02(4.9e-01) & \cellcolor{cyan!10}\textbf{-2.20e+02(4.4e-01)} \\
			F24 & \cellcolor{gray!8}4.50e+02(1.9e+01) & \cellcolor{gray!8}4.14e+02(1.9e+01) & \cellcolor{gray!8}3.98e+02(9.5e+00) & \cellcolor{gray!8}4.57e+02(1.3e+01) & \cellcolor{cyan!10}3.64e+02(1.3e+01) & \cellcolor{cyan!10}\textbf{3.53e+02(1.2e+01)} & \cellcolor{cyan!10}3.75e+02(1.5e+01) \\
			\midrule
			& \cellcolor{orange!7}\Large 24/0/0 & \cellcolor{orange!7}\Large 24/0/0 & \cellcolor{orange!7}\Large 21/1/2 & \cellcolor{orange!7}\Large 17/1/6 & \cellcolor{orange!7}\Large \checkmark & \cellcolor{orange!7}\Large - & \cellcolor{orange!7}\Large - \\
			& \cellcolor{orange!7}\Large 24/0/0 & \cellcolor{orange!7}\Large 24/0/0 & \cellcolor{orange!7}\Large 14/3/7 & \cellcolor{orange!7}\Large 15/1/8 & \cellcolor{orange!7}\Large - & \cellcolor{orange!7}\Large \checkmark & \cellcolor{orange!7}\Large - \\
			\multirow{-3}{*}{$+$/$\approx$/$-$} & \cellcolor{orange!7}\Large 24/0/0 & \cellcolor{orange!7}\Large 24/0/0 & \cellcolor{orange!7}\Large 11/4/9 & \cellcolor{orange!7}\Large 14/1/9 & \cellcolor{orange!7}\Large - & \cellcolor{orange!7}\Large - & \cellcolor{orange!7}\Large \checkmark \\
			\midrule
			\Large \textbf{Rank} & \cellcolor{orange!7}\Large 6.88 & \cellcolor{orange!7}\Large 5.50 & \cellcolor{orange!7}\Large 3.67 & \cellcolor{orange!7}\Large 4.04 & \cellcolor{orange!7}\Large \textbf{1.71} & \cellcolor{orange!7}\Large 2.71 & \cellcolor{orange!7}\Large 3.50 \\
			\bottomrule
		\end{tabular}
	}
			\caption{Comparison of objective values on BBOB benchmarks. Results are reported as Mean (Std) over 25 independent runs. The best mean values for each function are highlighted in bold.}
	\label{bigtab1}
\end{table*}
\subsection{Experimental Setup}
\label{subsec:experimental_setup}
To comprehensively evaluate AutoSG, we deploy it across three distinct optimization scenarios with varying dimensionalities and evaluation budgets:
\begin{itemize}[leftmargin=*]
	\item \textbf{High-Dimensional Expensive Optimization:} We utilize the established BBOB test suite \cite{hansen2021coco}, comprising 24 continuous problems. The dimensionality is set to $D=20$ with a budget of $300$ FEs. We select four representative baselines, namely \textbf{CMA-ES} \cite{hansen2016cma}, \textbf{TuRBO} \cite{eriksson2019scalable}, \textbf{HEBO} \cite{cowen2022hebo}, and \textbf{LLaMEA-BO} \cite{li2025llamea}. Specifically, for LLaMEA-BO, we directly evaluate \textbf{TREvol}, the optimal solver explicitly reported in the original publication for 20D problems.
	
	\item \textbf{Large-Scale Expensive Optimization:} We employ the CEC2013 test suite \cite{li2013benchmark} to assess AutoSG's scalability, consisting of 15 large-scale continuous optimization problems with dimensionalities ranging from $905\text{D}$ to $1000\text{D}$. The evaluation budget is linearly scaled to $11 \times D$ FEs. We select four representative baselines, namely \textbf{L2SMEA} \cite{l2smea}, \textbf{SAEA-RFS} \cite{saearfs}, \textbf{SADE-AMSS} \cite{9971764}, and \textbf{LSEO-S3} \cite{l2so}, alongside \textbf{LLaMEA-BO} for the ablation analysis.
	
	\item \textbf{Real-World Low-Dimensional Expensive Optimization:} We also test AutoSG on hyperparameter tuning (HPT) tasks derived from the Bayesmark test suite \cite{turner2019bayesmark}, involving 8 diverse datasets and 5 machine learning models (see Appendix \ref{app1}). These represent real-world problems with low dimensionalities ranging from 2D to 8D and a budget of only $30$ FEs. We select five representative baselines, namely \textbf{logEI} \cite{logei}, \textbf{CMA-ES}, \textbf{HEBO}, \textbf{TuRBO}, and \textbf{LLaMEA-BO} (specifically evaluating its optimal 5D and 10D solver: \textbf{ATRBO}).
\end{itemize}
To ensure a fair comparison, all baseline solvers are executed using the exact hyperparameter configurations recommended in their original publications. 
The specific hyperparameter settings and operational details for AutoSG are thoroughly documented in Appendix \ref{app2}. 
\subsection{Results}
\label{result}
\begin{table*}[t]
	\centering
	\resizebox{\textwidth}{!}{
		\setlength{\tabcolsep}{2pt} 
		\begin{tabular}{lccccccc}
			\toprule
			\multirow{2}{*}{\Large \textbf{Pro}} & \multicolumn{4}{c}{\cellcolor{gray!8}\Large \textbf{Baselines}} & \multicolumn{3}{c}{\cellcolor{cyan!10}\textbf{\Large Proposed AutoSG}} \\
			\cmidrule(lr){2-5} \cmidrule(lr){6-8}
			& \cellcolor{gray!8}L2SMEA & \cellcolor{gray!8}SAEA-RFS & \cellcolor{gray!8}SADE-AMSS & \cellcolor{gray!8}LSEO-S3 & 
			\cellcolor{cyan!10}$\mathcal{S}^*_{\text{CEC}, 1}$& \cellcolor{cyan!10}$\mathcal{S}^*_{\text{CEC}, 2}$ & \cellcolor{cyan!10}$\mathcal{S}^*_{\text{CEC}, 3}$ \\
			\midrule
			F1 & \cellcolor{gray!8}1.02e+10(7.6e+08) & \cellcolor{gray!8}7.47e+09(4.1e+09) & \cellcolor{gray!8}8.15e+09(6.3e+08) & \cellcolor{gray!8}7.39e+09(4.5e+08) & \cellcolor{cyan!10}3.52e+09(2.8e+08) & \cellcolor{cyan!10}\textbf{2.68e+09(3.0e+08)} & \cellcolor{cyan!10}4.73e+09(4.3e+08) \\
			F2 & \cellcolor{gray!8}2.36e+04(1.1e+03) & \cellcolor{gray!8}1.94e+04(1.1e+03) & \cellcolor{gray!8}2.26e+04(6.7e+02) & \cellcolor{gray!8}2.40e+04(4.1e+02) & \cellcolor{cyan!10}\textbf{1.76e+04(6.1e+02)} & \cellcolor{cyan!10}2.14e+04(1.4e+03) & \cellcolor{cyan!10}1.80e+04(3.3e+02) \\
			F3 & \cellcolor{gray!8}2.16e+01(8.9e-03) & \cellcolor{gray!8}2.09e+01(1.4e-02) & \cellcolor{gray!8}2.10e+01(2.0e-02) & \cellcolor{gray!8}2.11e+01(5.2e-03) & \cellcolor{cyan!10}2.10e+01(1.3e-02) & \cellcolor{cyan!10}2.10e+01(9.6e-03) & \cellcolor{cyan!10}\textbf{2.08e+01(1.1e-02)} \\
			F4 & \cellcolor{gray!8}1.10e+12(1.9e+11) & \cellcolor{gray!8}1.17e+12(4.6e+11) & \cellcolor{gray!8}6.86e+11(3.6e+11) & \cellcolor{gray!8}2.48e+11(3.5e+10) & \cellcolor{cyan!10}2.45e+11(5.1e+10) & \cellcolor{cyan!10}\textbf{2.09e+11(3.1e+10)} & \cellcolor{cyan!10}3.37e+11(9.4e+10) \\
			F5 & \cellcolor{gray!8}\textbf{4.07e+06(5.7e+05)} & \cellcolor{gray!8}2.29e+07(4.9e+06) & \cellcolor{gray!8}1.89e+07(2.8e+06) & \cellcolor{gray!8}1.91e+07(9.8e+05) & \cellcolor{cyan!10}1.77e+07(1.9e+06) & \cellcolor{cyan!10}1.80e+07(2.2e+06) & \cellcolor{cyan!10}1.77e+07(1.8e+06) \\
			F6 & \cellcolor{gray!8}1.07e+06(1.3e+03) & \cellcolor{gray!8}1.06e+06(3.4e+03) & \cellcolor{gray!8}1.06e+06(2.1e+03) & \cellcolor{gray!8}\textbf{1.06e+06(1.9e+03)} & \cellcolor{cyan!10}1.06e+06(2.7e+03) & \cellcolor{cyan!10}1.06e+06(2.2e+03) & \cellcolor{cyan!10}1.06e+06(2.1e+03) \\
			F7 & \cellcolor{gray!8}7.42e+10(2.7e+10) & \cellcolor{gray!8}7.63e+09(2.9e+09) & \cellcolor{gray!8}6.15e+09(2.4e+09) & \cellcolor{gray!8}2.75e+09(6.4e+08) & \cellcolor{cyan!10}\textbf{1.95e+09(6.3e+08)} & \cellcolor{cyan!10}2.89e+09(1.1e+09) & \cellcolor{cyan!10}2.18e+09(6.0e+08) \\
			F8 & \cellcolor{gray!8}1.21e+16(4.6e+15) & \cellcolor{gray!8}2.51e+16(1.9e+16) & \cellcolor{gray!8}2.21e+16(1.6e+16) & \cellcolor{gray!8}9.62e+15(1.2e+15) & \cellcolor{cyan!10}1.04e+16(3.4e+15) & \cellcolor{cyan!10}\textbf{7.82e+15(1.9e+15)} & \cellcolor{cyan!10}1.13e+16(4.4e+15) \\
			F9 & \cellcolor{gray!8}\textbf{5.03e+08(2.9e+07)} & \cellcolor{gray!8}1.71e+09(3.9e+08) & \cellcolor{gray!8}1.33e+09(1.9e+08) & \cellcolor{gray!8}1.47e+09(9.6e+07) & \cellcolor{cyan!10}1.29e+09(1.1e+08) & \cellcolor{cyan!10}1.29e+09(1.6e+08) & \cellcolor{cyan!10}1.34e+09(9.7e+07) \\
			F10 & \cellcolor{gray!8}9.52e+07(2.5e+05) & \cellcolor{gray!8}9.44e+07(6.1e+05) & \cellcolor{gray!8}9.46e+07(5.1e+05) & \cellcolor{gray!8}\textbf{9.39e+07(2.9e+05)} & \cellcolor{cyan!10}9.43e+07(3.9e+05) & \cellcolor{cyan!10}9.42e+07(4.0e+05) & \cellcolor{cyan!10}9.40e+07(4.1e+05) \\
			F11 & \cellcolor{gray!8}4.03e+11(8.8e+10) & \cellcolor{gray!8}1.02e+12(4.4e+11) & \cellcolor{gray!8}6.41e+11(2.9e+11) & \cellcolor{gray!8}3.53e+11(1.2e+11) & \cellcolor{cyan!10}2.44e+11(6.6e+10) & \cellcolor{cyan!10}\textbf{2.36e+11(5.2e+10)} & \cellcolor{cyan!10}3.42e+11(1.2e+11) \\
			F12 & \cellcolor{gray!8}1.15e+11(1.6e+10) & \cellcolor{gray!8}4.50e+11(4.0e+10) & \cellcolor{gray!8}1.25e+11(2.8e+10) & \cellcolor{gray!8}1.12e+11(9.1e+09) & \cellcolor{cyan!10}5.36e+10(6.3e+09) & \cellcolor{cyan!10}\textbf{2.22e+10(2.3e+09)} & \cellcolor{cyan!10}9.77e+10(1.2e+10) \\
			F13 & \cellcolor{gray!8}1.30e+11(4.6e+10) & \cellcolor{gray!8}9.57e+10(3.2e+10) & \cellcolor{gray!8}9.55e+10(2.0e+10) & \cellcolor{gray!8}2.85e+10(2.2e+09) & \cellcolor{cyan!10}\textbf{2.65e+10(3.6e+09)} & \cellcolor{cyan!10}4.10e+10(1.2e+10) & \cellcolor{cyan!10}2.89e+10(4.2e+09) \\
			F14 & \cellcolor{gray!8}\textbf{2.57e+11(8.6e+10)} & \cellcolor{gray!8}1.42e+12(4.2e+11) & \cellcolor{gray!8}1.17e+12(4.2e+11) & \cellcolor{gray!8}4.42e+11(7.3e+10) & \cellcolor{cyan!10}4.55e+11(9.7e+10) & \cellcolor{cyan!10}3.60e+11(1.1e+11) & \cellcolor{cyan!10}4.07e+11(9.3e+10) \\
			F15 & \cellcolor{gray!8}4.16e+11(1.5e+11) & \cellcolor{gray!8}4.11e+11(1.9e+12) & \cellcolor{gray!8}1.18e+08(3.5e+07) & \cellcolor{gray!8}6.88e+07(1.5e+07) & \cellcolor{cyan!10}5.27e+07(7.2e+06) & \cellcolor{cyan!10}7.81e+07(1.2e+07) & \cellcolor{cyan!10}\textbf{5.26e+07(6.7e+06)} \\
			\midrule
			& \cellcolor{orange!7}\Large 12/0/3 & \cellcolor{orange!7}\Large 13/1/1 & \cellcolor{orange!7}\Large 14/1/0 & \cellcolor{orange!7}\Large 9/5/1 & \cellcolor{orange!7}\Large \checkmark & \cellcolor{orange!7}\Large - & \cellcolor{orange!7}\Large - \\
			& \cellcolor{orange!7}\Large 12/0/3 & \cellcolor{orange!7}\Large 13/0/2 & \cellcolor{orange!7}\Large 12/2/1 & \cellcolor{orange!7}\Large 9/2/4 & \cellcolor{orange!7}\Large - & \cellcolor{orange!7}\Large \checkmark & \cellcolor{orange!7}\Large - \\
			\multirow{-3}{*}{$+$/$\approx$/$-$} & \cellcolor{orange!7}\Large 10/2/3 & \cellcolor{orange!7}\Large 15/0/0 & \cellcolor{orange!7}\Large 14/1/0 & \cellcolor{orange!7}\Large 8/5/2 & \cellcolor{orange!7}\Large - & \cellcolor{orange!7}\Large - & \cellcolor{orange!7}\Large \checkmark \\
			\midrule
			\Large \textbf{Rank} & \cellcolor{orange!7}\Large 5.27 & \cellcolor{orange!7}\Large 5.80 & \cellcolor{orange!7}\Large 5.33 & \cellcolor{orange!7}\Large 3.73 & \cellcolor{orange!7}\Large \textbf{2.40} & \cellcolor{orange!7}\Large 2.73 & \cellcolor{orange!7}\Large 2.73 \\
			\bottomrule
		\end{tabular}
	}
			\caption{Comparison of objective values on CEC2013 large-scale benchmarks. Results are reported as Mean (Std) over 25 independent runs. The best mean values for each function are highlighted in bold.}
	\label{bigtab2}
\end{table*}

\subsubsection{Results on the BBOB Test Suite}
\label{421}
We evaluate AutoSG and baseline solvers across the 24 continuous optimization problems in the BBOB test suite. To ensure statistical robustness, each solver is independently executed 25 times per instance. Table \ref{bigtab1} summarizes the average objective values and the average ranking of each solver.

To rigorously verify the stability and consistency of our generation framework, we independently execute AutoSG three times, denoting the resulting solvers as $\mathcal{S}^*_{\text{BBOB}, 1}$, $\mathcal{S}^*_{\text{BBOB}, 2}$, and $\mathcal{S}^*_{\text{BBOB}, 3}$. 
We evaluate the statistical significance between generated solvers and baselines using the Wilcoxon rank-sum test \cite{ranksum} at a 0.05 significance level. In Table \ref{bigtab1}, the symbols ``$+$'', ``$\approx$'', and ``$-$'' denote that the AutoSG-generated solver is significantly superior, equivalent, or inferior to the compared baseline, respectively. Finally, the overall statistical outcomes are summarized as aggregated (win/tie/loss) counts across all test instances.

As clearly evidenced by the results, the three independently generated solvers ($\mathcal{S}^*_{\text{BBOB}, 1}$, $\mathcal{S}^*_{\text{BBOB}, 2}$, and $\mathcal{S}^*_{\text{BBOB}, 3}$) remarkably secure the top three positions in the overall average ranking. Furthermore, the statistical test outcomes highlight their significant superiority over all compared baselines across the majority of test instances.

\subsubsection{Results on the CEC2013 Test Suite}
\label{422}
The comprehensive evaluation results comparing the three independently generated solvers ($\mathcal{S}^*_{\text{CEC}, 1}$, $\mathcal{S}^*_{\text{CEC}, 2}$, and $\mathcal{S}^*_{\text{CEC}, 3}$) against all baseline solvers on the CEC2013 test suite are summarized in Table \ref{bigtab2}. 
The experimental setup and statistical testing criteria remain identical to those detailed in Section \ref{421}.

Consistent with the exceptional performance observed on the BBOB benchmark, the three AutoSG-generated solvers remarkably secure the top three positions in the overall average ranking. Furthermore, the statistical test outcomes also demonstrate their significant superiority over all other evaluated baselines in tackling these highly challenging, large-scale optimization tasks.

\subsubsection{Results on HPT Tasks}
\label{subsubsec:hpt_results}

\begin{figure}[htbp]
	\centering
	\includegraphics[width=\columnwidth]{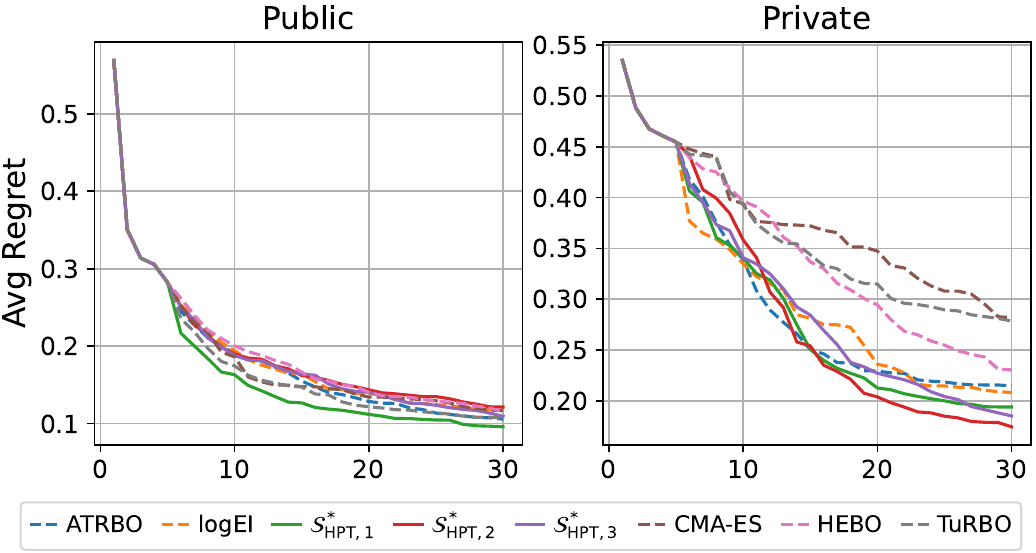} 
	\caption{Convergence curves of all solvers on the HPT tasks, averaged over 5 independent runs. The initial points are shared across all solvers.}
	\label{fig:hpt_convergence}
\end{figure}

The convergence curves comparing the three AutoSG-generated solvers ($\mathcal{S}^*_{\text{HPT}, 1}$, $\mathcal{S}^*_{\text{HPT}, 2}$, and $\mathcal{S}^*_{\text{HPT}, 3}$) against the baselines are illustrated in Figure \ref{fig:hpt_convergence}. 

As observed in the public datasets, while most evaluated solvers exhibit comparable performance with marginal differences, $\mathcal{S}^*_{\text{HPT}, 1}$ still achieves the best overall performance. More importantly, on the private datasets which are strictly unseen during the LLMs' pre-training phase---our generated solvers ($\mathcal{S}^*_{\text{HPT}, 1}$, $\mathcal{S}^*_{\text{HPT}, 2}$, and $\mathcal{S}^*_{\text{HPT}, 3}$) remarkably secure the top three positions. This robust performance on unobserved tasks strongly demonstrates the exceptional generalization capability of AutoSG.
\subsubsection{Ablation Experiments}
\noindent \textbf{Effectiveness of RAG-based Solver Generation.} We evaluate \textbf{AutoSG w/ RAG} (i.e., $\mathcal{S}_{init}$ derived via Eq.~\ref{eqinit}) against LLaMEA-BO and an ablation variant, \textbf{SA-LMCMAES} (i.e., $\tilde{\mathcal{S}}$ derived via Eq.~\ref{eqdllm}, using LLM self-recommendation without RAG), on the CEC2013 benchmark. As shown in Table~\ref{tab3}, lacking external knowledge, LLaMEA-BO defaults to Kriging-based solvers with prohibitive runtimes on large-scale tasks, yielding a 15/0/0 (win/tie/loss) record for AutoSG. Conversely, while the SA-LMCMAES solver generated via Eq.~\ref{eqdllm} is executable, factual hallucinations compromise its structural soundness, resulting in sub-SOTA performance. These results confirm that explicit RAG grounding is indispensable.

\begin{table}[htbp]
	\centering
	\setlength{\aboverulesep}{0pt}
	\setlength{\belowrulesep}{0pt}
	\resizebox{0.8\linewidth}{!}{
		\begin{tabular}{lc}
			\toprule
			 \textbf{Baselines} &  \textbf{AutoSG w/ RAG} \textit{\textbf{vs.}} \\
			\midrule
			 SA-LMCMAES & \cellcolor{orange!7} \textbf{12 / 0 / 3} \\
			 LLaMEA-BO  & \cellcolor{orange!7} \textbf{15 / 0 / 0} \\
			\bottomrule
		\end{tabular}
	}
	\caption{Statistical significance results ($+/{\approx}/-$) of AutoSG w/ RAG compared to baselines.}
		\label{tab3}
\end{table}
Furthermore, we verify the implementation faithfulness of $\mathcal{S}_{init}$. Since only the retrieved literature for the BBOB suite provides a public repository, we compare $\mathcal{S}_{init}$ directly against its official GitHub code for BBOB, and directly against $\mathcal{D}_{pdf}$ for the remaining two tasks. Specifically, by utilizing the web interface of Gemini 3 Pro with chain-of-thought reasoning to conduct rigorous cross-verification against the uploaded original source files, we observe a logical consistency rate achieving \textbf{95\%} across all three tasks. This strongly demonstrates the exceptional reproduction accuracy of our two-stage generation pipeline.

\begin{table}[htbp]
	\centering
	\setlength{\aboverulesep}{0pt}
	\setlength{\belowrulesep}{0pt}
	\resizebox{0.8\linewidth}{!}{
		\begin{tabular}{lc}
			\toprule
			\textbf{Baselines} & \textbf{AutoSG} \textit{\textbf{vs.}} \\
			\midrule
			AutoSG w/o SelfRefine & \cellcolor{orange!7}\textbf{9 / 2 / 4} \\
			AutoSG w/ C-M & \cellcolor{orange!7}\textbf{13 / 2 / 0} \\
			AutoSG w/ 2-step & \cellcolor{orange!7}\textbf{1 / 13 / 1} \\
			AutoSG w/ 10-step & \cellcolor{orange!7}\textbf{2 / 13 / 0} \\
			\bottomrule
		\end{tabular}
	}
	\caption{Statistical significance results ($+/{\approx}/-$) of AutoSG compared to its variants.}
	\label{tab:ablation_scec2}
\end{table}

\noindent \textbf{Effectiveness of the One-Step Self-Refine Operator.} We compare \textbf{AutoSG} ($\mathcal{S}^*_{\text{CEC}, 2}$) against \textbf{AutoSG w/o SelfRefine} (identical to AutoSG w/ RAG in Table \ref{tab3}), \textbf{AutoSG w/ C-M} (applying LLaMEA-BO's crossover-mutation to evolve 100 solvers from the top-2 RAG outputs), and multi-step variants (\textbf{AutoSG w/ 2-step} and \textbf{10-step}, which iteratively repeat the refinement and evaluation cycle on $\mathcal{S}^*$ for one and nine additional rounds). Results demonstrate the clear superiority of the one-step operator. Surprisingly, extensive iterations in AutoSG w/ C-M degrade performance compared to the variant without iterations. While viable for from-scratch generation, naive code crossover and mutation inherently destroy the established locally optimal heuristics generated via RAG, in stark contrast to the ``learn-and-improve'' paradigm of $\mathcal{P}_{refine}$ in AutoSG. Furthermore, the 2-step and 10-step variants yield no significant performance improvements. This indicates that performing excessive refinement iterations on RAG-grounded SOTA-level solvers is computationally redundant.

\begin{figure}[htbp]
	\centering
	\includegraphics[width=\columnwidth]{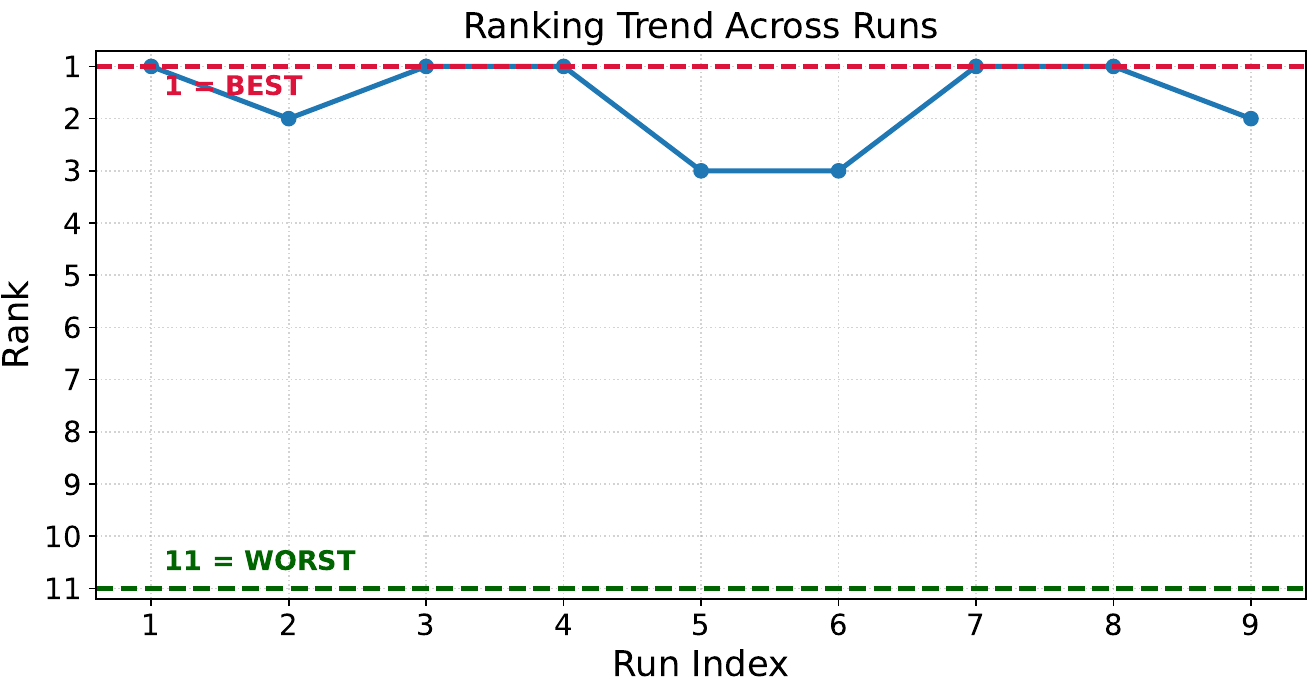} 
	\caption{The true ranking of $\mathcal{S}^*$ within $\mathcal{S}_{candidate}$ across 9 independent runs spanning three tasks.}
	\label{fig2}
\end{figure}

\noindent \textbf{Effectiveness of the LLM-as-a-Judge Evaluation.} Figure \ref{fig2} plots the true rankings of selected optimal solvers ($\mathcal{S}^*$) within their candidate pools ($\mathcal{S}_{candidate}$) across 9 independent runs (ground-truths averaged over 5 executions due to computational costs). Remarkably, our mechanism reliably identifies top-performing solver using only 63.64\% of exhaustive pair-wise matches, demonstrating significant efficiency gains.

Furthermore, Appendix \ref{app3} provides the intermediate artifact $\mathcal{D}_{pdf}$ generated by AutoSG and the executable source codes of the optimal solvers from three independent runs across all three tasks.

\section{Conclusions}
We introduced AutoSG, a fully automated LLM-driven workflow translating natural language prompts into customized solvers for expensive optimization. By grounding code generation in academic literature via RAG, AutoSG circumvents factual hallucinations. The one-step self-refinement operator introduces task-specific improvements while preserving locally optimal heuristics synthesized during generation. Finally, an Elo-based LLM-as-a-Judge mechanism rapidly establishes reliable global rankings. Extensive evaluations demonstrate AutoSG's exceptional versatility, significantly outperforming human-designed SOTA solvers and existing LLM-generated solvers.
\section*{Limitations}
As a fully LLM-driven workflow, AutoSG is inherently sensitive to external API stability. Currently, if any module fails to yield substantive content due to transient network timeouts or API invocation errors, the workflow halts and necessitates manual intervention to restart. To achieve true autonomy, future work will encapsulate AutoSG into a fully engineered, executable ``Skill'' driven by a self-reflective LLM agent. This agentic framework will be capable of autonomously diagnosing execution traces, troubleshooting errors, and recovering from runtime failures without any human oversight.

Beyond these engineering constraints, an important theoretical boundary of our framework warrants discussion. Since the generation engine is deeply rooted in retrieved academic literature, AutoSG naturally excels at interpolating and recombining existing domain knowledge. While highly effective, extending this capability to extrapolate entirely novel optimization paradigms completely outside the current human knowledge distribution remains an open challenge. Future research will explore advanced mathematical reasoning mechanisms that empower the model to discover structurally sound heuristics from first principles, thereby further expanding the innovation horizons beyond existing document retrieval.

\bibliography{main}

\clearpage
\appendix

\section{Detailed Descriptions of the Evaluated Optimization Tasks} \label{app1}

In this section, we provide detailed descriptions of the test suites and evaluation metrics adopted across the three distinct optimization tasks.

For the high-dimensional optimization task, we utilize the established BBOB test suite \cite{hansen2021coco}, which is widely recognized as a standard continuous optimization benchmark. It comprises 24 diverse problem instances, with the objective for each being to minimize its respective function value.

For the large-scale optimization task, we employ the CEC2013 test suite \cite{li2013benchmark}. This benchmark is extensively used to evaluate algorithms on 1000D search spaces and consists of 15 challenging large-scale continuous problems. Similar to the BBOB suite, the evaluation metric is the minimization of the objective function.

For the HPT task, we utilize the Bayesmark suite \cite{turner2019bayesmark} as a continuous HPT benchmark. We incorporate the 5 public datasets originally provided with the benchmark, alongside 3 private datasets sourced from \cite{liu2024large}. The evaluation spans across 5 representative machine learning models: RandomForest, SVM, DecisionTree, MLP, and AdaBoost. Consequently, each specific task is defined as a unique (dataset, model) pair. The primary scoring functions are accuracy for classification tasks and Mean Squared Error (MSE) for regression tasks.
For classification tasks (accuracy), where the optimal score is $s^{*}_{max}$, the normalized regret is formulated as:
\begin{equation}
	Regret_{cls}(t) = \frac{s^{*}_{max} - \max_{h \in H_t} (f(h))}{s^{*}_{max} - s^{*}_{min}}
\end{equation}
Conversely, for regression tasks (MSE), where the optimal score is $s^{*}_{min}$, the normalized regret is formulated as:
\begin{equation}
	Regret_{reg}(t) = \frac{\min_{h \in H_t} (f(h)) - s^{*}_{min}}{s^{*}_{max} - s^{*}_{min}}
\end{equation}
where $H_t$ denotes the set of hyperparameter configurations evaluated up to trial $t$. Therefore, regardless of the task type, a regret value of 0 consistently indicates hitting the theoretical optimal performance. 
Furthermore, we strictly follow the search spaces designated in the Bayesmark benchmark, specifying the hyperparameter types, transform spaces, and boundary ranges (lower and upper bounds). These specified search spaces are uniformly applied across all baseline solvers, with necessary space transformations performed prior to the optimization process. The detailed search space for each machine learning model is summarized below in the format \texttt{\{hyperparameter name: [space transform, lower bound, upper bound]\}}:

\begin{itemize}[leftmargin=*]
	\item \textbf{SVM [3D]:} \texttt{\{C: [log, 1, 1e3], $\gamma$: [log, 1e-4, 1e-3], tolerance: [log, 1e-5, 1e-1]\}}
	\item \textbf{DecisionTree [6D]:} \texttt{\{max\_depth: [linear, 1, 15], min\_samples\_split: [logit, 0.01, 0.99], min\_samples\_leaf: [logit, 0.01, 0.49], min\_weight\_fraction\_leaf: [logit, 0.01, 0.49], max\_features: [logit, 0.01, 0.99], min\_impurity\_decrease: [linear, 0.0, 0.5]\}}
	\item \textbf{RandomForest [6D]:} \texttt{\{max\_depth: [linear, 1, 15], min\_samples\_split: [logit, 0.01, 0.99], min\_samples\_leaf: [logit, 0.01, 0.49], min\_weight\_fraction\_leaf: [logit, 0.01, 0.49], max\_features: [logit, 0.01, 0.99], min\_impurity\_decrease: [linear, 0.0, 0.5]\}}
	\item \textbf{MLP [8D]:} \texttt{\{hidden\_layer\_sizes: [linear, 50, 200], alpha: [log, 1e-5, 1e1], batch\_size: [linear, 10, 250], learning\_rate\_init: [log, 1e-5, 1e-1], power\_t: [logit, 0.1, 0.9], tol: [log, 1e-5, 1e-1], momentum: [logit, 0.001, 0.999], validation\_fraction: [logit, 0.1, 0.9]\}}
	\item \textbf{AdaBoost [2D]:} \texttt{\{n\_estimators: [linear, 10, 100], learning\_rate: [log, 1e-4, 1e1]\}}
\end{itemize}

\section{Implementation Details of AutoSG} \label{app2}
To ensure full reproducibility of our proposed framework, we provide the comprehensive hyperparameter configurations used across all modules of AutoSG in Table \ref{tab:hyperparameters}. 
Specifically, the parameter configurations governing the Elo rating system, including the initial rating $R_i^{(0)}$, the initial rating deviation $rd_i^{(0)}$, the pruning threshold $\Delta_{max}$, and the scaling factors for $K_i$, are established strictly in accordance with well-established literature standards \cite{zhang2026uda,sendagorta2021elo}.
Furthermore, all prompt templates ($\mathcal{P}$) utilized throughout the AutoSG workflow are presented in Figures~\ref{figptask}-\ref{fig:prompt_algorithm_comparison}. Additionally, the specific definitions of the components within the task-specific prompt $\mathcal{P}_{task}$ for various optimization tasks are detailed in Table~\ref{ptaskt}.

\begin{table*}[htbp]
	\centering
	\renewcommand{\arraystretch}{1.2}
	\resizebox{\linewidth}{!}{
		\begin{tabular}{llc}
			\toprule
			\textbf{Symbol} & \textbf{Description} & \textbf{Value} \\
			\midrule
			\multicolumn{3}{c}{\textit{Base Configuration}} \\
			\midrule
			- & Underlying LLM & \texttt{Claude Opus 4.6} \\
			\midrule
			\multicolumn{3}{c}{\textit{RAG-based Solver Generation (cf. Section \ref{sec32})}} \\
			\midrule
			$K$ & Number of rewritten search queries & 8 \\
			$M_{OA}$ & Top recall limit per query for OpenAlex & 30 \\
			$M_{ax}$ & Top recall limit per query for arXiv & 10 \\
			$N_{OA}$ & OpenAlex candidate pool size after deduplication & 25 \\
			$N_{ax}$ & arXiv candidate pool size after deduplication & 15 \\
			$N$ & Total candidate papers for LLM reranking & 40 \\
			\midrule
			\multicolumn{3}{c}{\textit{One-Step Solver Self-Refinement (cf. Section \ref{sec33})}} \\
			\midrule
			$C$ & Number of refined candidate solvers & 10 \\
			$|\mathcal{S}_{candidate}|$ & Total size of the evaluation pool & 11 (1 initial + 10 refined) \\
			\midrule
			\multicolumn{3}{c}{\textit{Elo-based LLM-as-a-Judge Evaluation (cf. Section \ref{sec34})}} \\
			\midrule
			$R_i^{(0)}$ & Initial Elo rating & 1500 \\
			$rd_i^{(0)}$ & Initial rating deviation & 350 \\
			$\Delta_{max}$ & Rating difference threshold for match pruning & 400 \\
			$T_{init}$ & Minimum matches per solver in Phase 1 & 3 \\
			$T_{total}$ & Target total matches per solver & 6 \\
			$\lambda$ & Balancing coefficient in Phase 2 heuristic & 10 \\
			\bottomrule
		\end{tabular}
	}
	\caption{Hyperparameter configurations across different modules of the AutoSG workflow.}
	\label{tab:hyperparameters}
\end{table*}

\begin{figure*}[t]
	\begin{tcblisting}{
			colback=gray!8, 
			colframe=gray!70, 
			width=\textwidth, 
			size=small,
			listing only, 
			listing options={
				basicstyle=\ttfamily\small, 
				breaklines=true, 
				breakatwhitespace=true,
				columns=fullflexible,
				escapechar=|,
				literate={—}{{-}}1 {–}{{-}}1
			}
		}
You are a highly skilled computer scientist in the field of expensive black-box optimization, Bayesian optimization, surrogate-assisted evolutionary computation, and modern metaheuristic algorithms. Your task is to design novel optimization algorithms to solve expensive black-box optimization problems

The algorithm is evaluated on the |\textcolor{blue}{\{$\mathcal{T}_{desc}$ in $\mathcal{P}_{task}$\}}|. Your task is to write the optimization algorithm in Python code. The code should contain an `__init__(self, budget, dim)` function and the function `__call__(self, func)`, which should *minimize* the expensive black-box function `func` using `self.budget` function evaluations.
The func() can only be called as many times as the budget allows, not more. Each of the optimization functions has a search space of |\textcolor{blue}{\{$\mathcal{X}$ in $\mathcal{P}_{task}$\}}| (bounds are set in self.bounds). **The algorithm must be specifically designed and optimized to solve |\textcolor{blue}{\{$\mathcal{T}_{desc}$ in $\mathcal{P}_{task}$\}}| with a strict budget of |\textcolor{blue}{\{$B$ in $\mathcal{P}_{task}$\}}| function evaluations.** The algorithm MUST use a surrogate model to approximate the expensive function and guide the search, thereby reducing the number of real function evaluations.
As an expert of numpy, scipy, scikit-learn, torch, gpytorch, you are allowed to use these libraries. Do not use any other libraries unless they cannot be replaced by the above libraries. Do not remove the comments from the code.
Name the class based on the characteristics of the algorithm.

Design an excellent optimization algorithm with strong performance to solve this task. While novel designs are welcome, well-established and widely verified algorithms are also fully acceptable. The algorithm can use any suitable approach including but not limited to:
- Evolutionary algorithms
- Bayesian optimization
- Surrogate-assisted methods
- Model-based optimization
- Metaheuristic algorithms
- Hybrid approaches

Key design considerations:
- How to efficiently use the limited |\textcolor{blue}{\{$B$\}}| budget 
- Balancing exploration and exploitation in search spaces
- Handling |\textcolor{blue}{\{$d$\}}|-D dimensionality
	\end{tcblisting}

	\caption{Content for $\mathcal{P}_{task}$.}
	\label{figptask}
\end{figure*}

\begin{table*}[t]
	\centering
	\large
	\begin{tabularx}{\textwidth}{@{} >{\bfseries}l c c >{\raggedright\arraybackslash}X >{\raggedright\arraybackslash}X @{}}
		\toprule
		Task Suite & $\bm{d}$ & $\bm{B}$ & $\bm{\mathcal{T}_{desc}}$ & $\bm{\mathcal{X}}$ \\
		\midrule
		
		BBOB & 20D & 300 FEs & 
		20-dimensional (20D) high-dimensional expensive problems & 
		between -5.0 (lower bound) and 5.0 (upper bound) \\
		\addlinespace
		
		CEC2013 & 1000D & $11d$ FEs & 
		1000-dimensional (1000D) large-scale expensive problems & 
		[-100.0, 100.0], or [-5.0, 5.0], or [-32.0, 32.0] per dimension \\
		\addlinespace
		
		Bayesmark & 2--8D & 30 FEs & 
		low-dimensional expensive problems with dimension $d$ in the range 2 to 8 (2--8D) & 
		heterogeneous and problem-dependent bounds, where each decision variable has distinct scale limits (e.g., [1, 1e3] vs. [0.01, 0.49]) instead of a universal fixed box \\
		
		\bottomrule
	\end{tabularx}
	\caption{Definitions of $\mathcal{P}_{task}$ components across different optimization tasks.}
	\label{ptaskt}
\end{table*}

\begin{figure*}[t]
	\begin{tcblisting}{
			colback=gray!8, 
			colframe=gray!70, 
			width=\textwidth, 
			size=small,
			listing only, 
			listing options={
				basicstyle=\ttfamily\small, 
				breaklines=true, 
				breakatwhitespace=true,
				columns=fullflexible,
				escapechar=|,
				literate={—}{{-}}1 {–}{{-}}1
			}
		}
You are a senior researcher specializing in academic literature retrieval. You are proficient in information retrieval, keyword extraction, and query formulation for academic search engines such as OpenAlex, arXiv, Google Scholar, IEEE Xplore, Scopus, and Web of Science. Your goal is to help the user design comprehensive and precise search queries that maximize recall of relevant state-of-the-art papers while minimizing irrelevant results.
	
I am conducting an academic literature search. Below is the problem description / prompt file that defines what I need to solve. Please read it carefully and help me summarize the best keywords and search queries.
	
Problem Description:
|\textcolor{blue}{\{$\mathcal{P}_{task}$\}}|
	
Your Task:
Based on the problem description above, generate exactly 8 search query strings that can be directly used as queries for OpenAlex and arXiv academic search APIs. The 8 queries MUST follow this structure:
	
- Query 1: Problem name query. Use the specific problem/application name from the description.
- Queries 2-4: Broad queries (high recall). Use short, general keyword combinations (3-5 words) that match the core research area. These ensure important papers are not missed.
- Queries 5-8: Precise queries (high precision). Use more specific keyword combinations (5-8 words) that narrow down to the exact problem setting.

Requirements:
- Queries should be concise keyword combinations (not full sentences).
- Use English academic terminology.
- ONLY extract keywords that describe the problem setting and research direction (e.g., problem type, scale, budget constraint, optimization category).
- Do NOT include specific algorithm names, specific surrogate model types (e.g., RBF, Gaussian process), specific techniques (e.g., random embedding, Latin hypercube), or any concrete solution methods — those are implementation details, not search keywords for finding relevant literature.

Output ONLY the following JSON block, nothing else:
{"search_queries": ["problem_name", "broad1", "broad2", "broad3", "precise1", "precise2", "precise3", "precise4"]}
\end{tcblisting}
\captionof{figure}{Content for $\mathcal{P}_{rewrite}$.}
\label{fig:prompt_literature_search}
\end{figure*}

\begin{figure*}[t]
	\begin{tcblisting}{
			colback=gray!8, 
			colframe=gray!70, 
			width=\textwidth, 
			size=small,
			listing only, 
			listing options={
				basicstyle=\ttfamily\small, 
				breaklines=true, 
				breakatwhitespace=true,
				columns=fullflexible,
				escapechar=|,
				literate={—}{{-}}1 {–}{{-}}1
			}
		}
You are a highly skilled computer scientist in the field of expensive black-box optimization, Bayesian optimization, surrogate-assisted evolutionary computation, and modern metaheuristic algorithms. You stay current with the latest publications from top-tier venues. Your task is to identify the most effective state-of-the-art optimization algorithms for the given setting, prioritizing recent advances and empirical performance over traditional methods.

I need you to analyze recent literature and identify the SINGLE BEST algorithm for the following setting:
**Problem Setting:**
- Dimensionality: |\textcolor{blue}{\{$d$ in $\mathcal{P}_{task}$\}}|
- Function Evaluation Budget:  |\textcolor{blue}{\{$B$ in $\mathcal{P}_{task}$\}}|
- Problem Type: |\textcolor{blue}{\{$\mathcal{T}_{desc}$ in $\mathcal{P}_{task}$\}}|
- Search space / bounds: |\textcolor{blue}{\{$\mathcal{X}$ in $\mathcal{P}_{task}$\}}|

Below is a curated collection of recent academic papers retrieved from OpenAlex and arXiv. Please read them carefully.

|\textcolor{blue}{\{$meta_1, \dots, meta_N$\}}|

**Your Task:**
**IMPORTANT: Do NOT recommend survey, review, or benchmarking/comparison papers — you must recommend the ORIGINAL paper that proposes the algorithm, with sufficient algorithmic detail (pseudocode, equations) for implementation.**

**Important: Prioritize papers published in top-tier venues**, including but not limited to:
- IEEE Transactions (e.g., IEEE TEVC, IEEE TCYB, IEEE TNNLS, IEEE TAI, IEEE TSMC)
- ACM conferences and journals
- Top AI/ML conferences (NeurIPS, ICML, ICLR, AAAI, IJCAI, GECCO, CEC, PPSN)

Papers from these venues should be given significantly higher weight than workshop papers, arXiv-only preprints, or low-tier venues.

Based on ALL the above sources and your own expertise, please provide:

1. **Top-1 Candidate Algorithm**: Name the single most promising algorithm for the **|\textcolor{blue}{\{$d$\}}| / |\textcolor{blue}{\{$B$\}}|** setting as described above. Cite the specific paper and its publication venue, and explain its key mechanism.

2. **Comparative Analysis**: Create a comparison table with columns: Algorithm Name, Paper Reference, Venue, Key Mechanism, Strengths for **|\textcolor{blue}{\{$d$\}}| / |\textcolor{blue}{\{$B$\}}|**, Weaknesses/Limitations, Estimated Suitability (1-10). 

3. **Final Recommendation — Top 1**: Select the ONE BEST algorithm and justify your choice in detail. Consider:
- Effectiveness in **|\textcolor{blue}{\{$d$\}}|** expensive optimization
- Efficiency with a very limited **fixed** budget of **|\textcolor{blue}{\{$B$\}}|**
- Availability of implementation details in the paper
- Publication venue quality (prefer IEEE Trans / top conferences)

4. **CRITICAL — Structured Output**: At the very end of your response, output the following JSON block (this will be parsed programmatically). Use the EXACT paper title as it appears in the literature above:

```json
{
	"top1": [
	{
		"rank": 1,
		"algorithm_name": "<short algorithm name/abbreviation>",
		"paper_title": "<exact full paper title>",
		"venue": "<publication venue>",
		"year": <year>,
		"reason": "<one-sentence justification>"
	}
	]
}
\end{tcblisting}
\captionof{figure}{Content for $\mathcal{P}_{rerank}$.}
\label{fig:prompt_algorithm_recommendation}
\end{figure*}

\begin{figure*}[t]
	\begin{tcblisting}{
			colback=gray!8, 
			colframe=gray!70, 
			width=\textwidth, 
			size=small,
			listing only, 
			listing options={
				basicstyle=\ttfamily\small, 
				breaklines=true, 
				breakatwhitespace=true,
				columns=fullflexible,
				escapechar=|,
				literate={—}{{-}}1 {–}{{-}}1
			}
		}
You are a highly skilled computer scientist in the field of expensive black-box optimization, Bayesian optimization, surrogate-assisted evolutionary computation, and modern metaheuristic algorithms. You stay current with the latest publications from top-tier venues. Your task is to faithfully implement optimization algorithms based on their original paper content.

You must implement the algorithm from the paper "|\textcolor{blue}{\{paper\_title from $meta$\}}|" (|\textcolor{blue}{\{venue from $meta$\}}|, |\textcolor{blue}{\{year from $meta$\}}|) 

**Problem Setting:**
- Dimensionality: |\textcolor{blue}{\{$d$ in $\mathcal{P}_{task}$\}}|
- Function Evaluation Budget: |\textcolor{blue}{\{$B$ in $\mathcal{P}_{task}$\}}|
- Problem Type: |\textcolor{blue}{\{$\mathcal{T}_{desc}$ in $\mathcal{P}_{task}$\}}|
- Search space / bounds: |\textcolor{blue}{\{$\mathcal{X}$ in $\mathcal{P}_{task}$\}}|

--- FULL PAPER CONTENT ---

|\textcolor{blue}{\{$\mathcal{D}_{pdf}$\}}|

--- END OF PAPER ---

**Implementation Requirements:**

Based on the FULL PAPER CONTENT above, provide a HIGH-QUALITY, FEATURE-COMPLETE implementation:
- Read the paper carefully and implement ALL key components and mechanisms described
- Do NOT oversimplify by replacing sophisticated strategies with basic alternatives
- Follow the algorithm pseudocode and equations in the paper as closely as possible
- Aim for a production-quality implementation that faithfully represents the algorithm

Please:
1. First, briefly summarize the key algorithmic components you identified from the paper
2. Then provide the complete Python implementation following the code template below
3. |\textcolor{blue}{$\mathcal{P}_{imp}$: }| For each key component (including formula, operator, parameter) that the paper emphasizes as critical or contributing to performance, add a trailing comment `# **IMPORTANT COMPONENT**` at the corresponding function or code block. This helps identify which parts are the paper's core contributions vs. standard boilerplate.

|\textcolor{blue}{\{$\mathcal{T}_{code}$\}}|
\end{tcblisting}
\captionof{figure}{Content for $\mathcal{P}_{G1}$}
\label{fig:prompt_algorithm_implementation}
\end{figure*}

\begin{figure*}[t]
	\begin{tcblisting}{
			colback=gray!8, 
			colframe=gray!70, 
			width=\textwidth, 
			size=small,
			listing only, 
			listing options={
				basicstyle=\ttfamily\small, 
				breaklines=true, 
				breakatwhitespace=true,
				columns=fullflexible,
				escapechar=|,
				literate={—}{{-}}1 {–}{{-}}1
			}
		}
```python
from collections.abc import Callable
import numpy as np
import torch
import gpytorch
# Import any additional libraries you need for your algorithm design

class <AlgorithmName>:
def __init__(self, budget: int, dim: int):
self.budget = budget
self.dim = dim
# bounds: shape (2, dim); bounds[0]=lower, bounds[1]=upper per coordinate (often heterogeneous / problem-dependent).
self.bounds = np.array([[-5.0] * dim, [5.0] * dim])

# Data storage for evaluated points (if needed)
self.X: np.ndarray = None  # shape (n_points, n_dims) - stores evaluated points
self.y: np.ndarray = None  # shape (n_points, 1) - stores evaluated values
self.n_evals = 0  # Number of function evaluations

# Algorithm-specific parameters
# Define your strategy parameters here

# GPU/Device configuration (REQUIRED if using torch/gpytorch)
self.device = torch.device("cuda" if torch.cuda.is_available() else "cpu")
self.dtype = torch.float64  # or torch.float32 for faster computation

# Do not add any other arguments without a default value

def __call__(self, func: Callable[[np.ndarray], np.float64]) -> tuple[np.float64, np.array]:
# Main optimization loop
# func: takes array of shape (n_dims,) and returns np.float64
# !!! CRITICAL: Monitor self.n_evals and ensure it never exceeds self.budget !!!
# Return a tuple (best_y, best_x) where:
#   - best_y: best objective value found (scalar)
#   - best_x: best solution found (1D numpy array of shape (dim,))

# Implement your algorithm here

return best_y, best_x
```

**Code Requirements:**
- The code should contain an `__init__(self, budget, dim)` function and the function `__call__(self, func)`, which should *minimize* the expensive black-box function `func` using `self.budget` function evaluations.
- The func() can only be called as many times as the budget allows, not more.
- Use `self.bounds` with shape (2, dim) for per-coordinate lower/upper bounds; they are problem-dependent and may be heterogeneous (e.g., [1, 1e3], [0.01, 0.49], [10, 100] across variables). Do not hard-code a single global interval for all problems unless that is the evaluated protocol.
- You are allowed to use numpy, scipy, scikit-learn, torch, gpytorch libraries.
- Name the class based on the algorithm name or its characteristics.
- Keep the code structure and comments from the template.
- **GPU acceleration support is RECOMMENDED when using torch/gpytorch**. Use `torch.device("cuda" if torch.cuda.is_available() else "cpu")` to automatically detect and use GPU when available.
\end{tcblisting}
\captionof{figure}{Content for $\mathcal{T}_{code}$}
\label{fig:prompt_algorithm_implementa}
\end{figure*}

\begin{figure*}[t]
	\begin{tcblisting}{
			colback=gray!8, 
			colframe=gray!70, 
			width=\textwidth, 
			size=small,
			listing only, 
			listing options={
				basicstyle=\ttfamily\small, 
				breaklines=true, 
				breakatwhitespace=true,
				columns=fullflexible,
				escapechar=|,
				literate={—}{{-}}1 {–}{{-}}1
			}
		}
You are a meticulous algorithm verification expert. Your job is to compare a Python implementation against its original paper and identify ALL discrepancies. You must fix every deviation from the paper's pseudocode, equations, and algorithm descriptions. Do NOT introduce your own knowledge — only follow what the paper explicitly states.

Below is a Python implementation of the algorithm "|\textcolor{blue}{\{algo\_name for $\mathcal{S}_{pre}$ }\}|" from the paper "|\textcolor{blue}{\{paper\_title from $meta$\}}|".

Your task: carefully compare this implementation against the ORIGINAL PAPER and fix ALL discrepancies.

--- FULL PAPER CONTENT (for reference) ---

|\textcolor{blue}{\{$\mathcal{D}_{pdf}$\}}|

--- END OF PAPER ---

--- CURRENT IMPLEMENTATION (Round 1 output) ---

```python
|\textcolor{blue}{\{$\mathcal{S}_{pre}$\}}|
```

--- END OF IMPLEMENTATION ---

**Review Checklist — go through each item and fix if needed:**

1. **Algorithm pseudocode**: Compare EVERY line of the paper's Algorithm/pseudocode with the code. Are all steps implemented? Is the order correct? Are any steps missing or extra?

2. **Equations**: Check EVERY numbered equation in the paper. Is each equation translated to code correctly?

3. **Parameters**: Are all algorithm parameters set to the values specified in the paper?

4. **Data flow**: After environmental selection or any selection step, are the selected individuals' associated data correctly tracked and assigned?

5. **Boundary handling**: Does the code handle boundary constraints the same way as described in the paper?

6. **Training sample selection**: How are training samples selected for the surrogate model? Does it match the paper?

**Output Requirements:**
- First, list ALL discrepancies you found (numbered list), with the specific paper reference (equation number, algorithm line, section) for each
- Then provide the COMPLETE corrected Python implementation (not just patches — the full code)
- The corrected code must follow the same class template structure (__init__ with budget/dim, __call__ with func)
- Keep the same class name: |\textcolor{blue}{\{algo\_name for $\mathcal{S}_{pre}$}\}|
- For each key component (including formula, operator, parameter) that the paper emphasizes as critical or contributing to performance, add a trailing comment `# **IMPORTANT COMPONENT**` at the corresponding function or code block. This helps identify which parts are the paper's core contributions vs. standard boilerplate.
\end{tcblisting}
\captionof{figure}{Content for $\mathcal{P}_{G2}$}
\label{fig:prompt_algorithm_verification}
\end{figure*}

\begin{figure*}[t]
	\begin{tcblisting}{
			colback=gray!8, 
			colframe=gray!70, 
			width=\textwidth, 
			size=small,
			listing only, 
			listing options={
				basicstyle=\ttfamily\small, 
				breaklines=true, 
				breakatwhitespace=true,
				columns=fullflexible,
				escapechar=|,
				literate={—}{{-}}1 {–}{{-}}1
			}
		}
Analyze the following optimization algorithm implementation and provide a concise technical description (3-5 sentences). 
Focus on: (1) the algorithm's name and origin (paper/venue if identifiable from code comments), (2) its core strategy and key components, (3) what makes it suitable for |\textcolor{blue}{\{$\mathcal{T}_{desc}$ in $\mathcal{P}_{task}$\}}|.
	
Do NOT include any code. Output ONLY the description text, nothing else.
	
"""python
|\textcolor{blue}{\{$\mathcal{S}_{init}$}\}|
"""
	\end{tcblisting}
	\captionof{figure}{Content for $\mathcal{P}_{reverse}$}
	\label{fig:prompt_new_algorithm_design1}
\end{figure*}

\begin{figure*}[t]
	\begin{tcblisting}{
			colback=gray!8, 
			colframe=gray!70, 
			width=\textwidth, 
			size=small,
			listing only, 
			listing options={
				basicstyle=\ttfamily\small, 
				breaklines=true, 
				breakatwhitespace=true,
				columns=fullflexible,
				escapechar=|,
				literate={—}{{-}}1 {–}{{-}}1
			}
		}
|\textcolor{blue}{\{$\mathcal{P}_{task}$\}}|

**IMPORTANT Performance Warning:**
- Do NOT use `scipy.optimize.minimize` (e.g., L-BFGS-B) to optimize acquisition functions. It uses numerical gradients which are extremely slow.

Give it a concise but comprehensive key-word-style description with the main ideas and justify your decision about the algorithm design.

Study the high-performing algorithm carefully and **learn** the key problem-solving strategies, design patterns, and algorithmic ideas that make them effective.
Based on what you learn, design a **new** algorithm. 

**IMPORTANT COMPONENT preservation:** In the parent code above, any line or block annotated with `# **IMPORTANT COMPONENT**` marks mechanisms the original paper emphasized as critical. **Change these as little as possible** when you design the new algorithm: keep the same ideas, structure, and behavior unless a minimal edit is strictly required for integration or correctness. Do not replace them with generic shortcuts. If you reuse or adapt such a block, retain the `# **IMPORTANT COMPONENT**` comment on the corresponding code.

##  |\textcolor{blue}{\{algo\_name for $\mathcal{S}_{init}$}\}|
##  |\textcolor{blue}{\{$\mathcal{D}_{init}$}\}|

With code:
```python
"""
SOTA Algorithm for |\textcolor{blue}{\{$d$\}}| / |\textcolor{blue}{\{$B$\}}| Setting
Algorithm:  |\textcolor{blue}{\{algo\_name for $\mathcal{S}_{init}$}\}|
Paper: |\textcolor{blue}{\{paper\_title from $meta$\}}|
Venue: |\textcolor{blue}{\{venue from $meta$\}}|
Implementation source: Full paper PDF

Problem Setting:
- Dimensionality: |\textcolor{blue}{\{$d$ in $\mathcal{P}_{task}$\}}|
- Function Evaluation Budget: |\textcolor{blue}{\{$B$ in $\mathcal{P}_{task}$\}}|
- Problem Type: |\textcolor{blue}{\{$\mathcal{T}_{desc}$ in $\mathcal{P}_{task}$\}}|
- Search space / bounds: |\textcolor{blue}{\{$\mathcal{X}$ in $\mathcal{P}_{task}$\}}|
"""

|\{\textcolor{blue}{$\mathcal{S}_{init}$}\}|
```

A code structure guide is as follows and keep the comments from the guide when generating the code.

|\textcolor{blue}{\{$\mathcal{T}_{code}$\}}|

Give the response in the format:
# Analysis
<Analyze the specific difficulties of the |\textcolor{blue}{\{$d$\}}| / |\textcolor{blue}{\{$B$\}}| setting and how this impacts algorithm choice>
# Description 
<description>
# Justification 
<justification for the key components of the algorithm or the changes made>
# Code 
<code>
\end{tcblisting}
\captionof{figure}{Content for $\mathcal{P}_{refine}$}
\label{fig:prompt_new_algorithm_design}
\end{figure*}

\begin{figure*}[t]
\begin{tcblisting}{
		colback=gray!8, 
		colframe=gray!70, 
		width=\textwidth, 
		size=small,
		listing only, 
		listing options={
			basicstyle=\ttfamily\small, 
			breaklines=true, 
			breakatwhitespace=true,
			columns=fullflexible,
			escapechar=|,
			literate={—}{{-}}1 {–}{{-}}1
		}
	}
You are a highly skilled computer scientist in the field of expensive black-box optimization, Bayesian optimization, surrogate-assisted evolutionary computation, and modern metaheuristic algorithms.

Your task is to rigorously evaluate and compare two optimization algorithm implementations (Algorithm A and Algorithm B) designed to solve |\textcolor{blue}{\{$\mathcal{T}_{desc}$ in $\mathcal{P}_{task}$\}}| problems.

### The Problem Context
The algorithms must minimize expensive black-box objectives with a search space of |\textcolor{blue}{\{$\mathcal{X}$ in $\mathcal{P}_{task}$\}}|. **The target regime is |\textcolor{blue}{\{$d$ in $\mathcal{P}_{task}$\}}| with a strict budget of |\textcolor{blue}{\{$B$ in $\mathcal{P}_{task}$\}}| per run**. Implementations must be in Python, using libraries like numpy, scipy, scikit-learn, torch, or gpytorch.

### Evaluation Criteria

1.  **Algorithm Design & Strategy:**
-   Does the algorithm use an appropriate optimization strategy given **very scarce data** (|\textcolor{blue}{\{$B$\}}|) and **dimensionality (|\textcolor{blue}{\{$d$\}}|)**?
-   How efficiently does the algorithm use the limited **|\textcolor{blue}{\{$B$\}}| function evaluations**?

2.  **Exploration-Exploitation Balance:**
-   How does the algorithm select candidates for evaluation?
-   Does the design effectively balance exploration and exploitation?

3.  **Implementation Correctness:**
-   Does the code strictly respect the **|\textcolor{blue}{\{$B$\}}|** limit?
-   Is the code logic sound? Are there shape mismatches, data leakage, or initialization errors?

4.  **Robustness:**
-   Is the design appropriate for the **|\textcolor{blue}{\{$d$\}}| / |\textcolor{blue}{\{$B$\}}|** expensive setting?

### Judgment Logic

1.  **Focus on Design Quality:** Evaluate solely the algorithm's optimization quality and design soundness. Do not consider computational speed or execution time.

2.  **Anti-bias:** Do not treat longer code, more complex class names, or stacking more modules as stronger evidence of a better algorithm.

### Output Requirement

Return a JSON object (no markdown code blocks):
{
	"analysis": "A detailed technical analysis comparing Algorithm A and B, discussing their optimization strategies, sample efficiency, and suitability for the |\textcolor{blue}{\{$d$\}}| / |\textcolor{blue}{\{$B$\}}| expensive setting.",
	"winner": "Algorithm A" or "Algorithm B"
}
|\textcolor{blue}{\{algorithm A\}}|
|\textcolor{blue}{\{algorithm B\}}|
\end{tcblisting}
\captionof{figure}{Content for $\mathcal{P}_{judge}$}
\label{fig:prompt_algorithm_comparison}
\end{figure*}

\section{Execution Results of AutoSG} \label{app3}
In this section, we detail the intermediate artifacts, specifically the retrieved academic literature ($\mathcal{D}_{pdf}$), and provide the source codes of the final generated solvers across the three evaluated optimization tasks.
Table \ref{tab:retrieved_papers} summarizes the exact SOTA-level papers autonomously retrieved by AutoSG's RAG module for each respective task domain. 

\begin{table*}[htbp]
	\centering
	\renewcommand{\arraystretch}{1.5} 
	\Large 
	\begin{tabularx}{\textwidth}{@{} 
			>{\raggedright\arraybackslash\hsize=0.7\hsize}X 
			>{\raggedright\arraybackslash\hsize=1.3\hsize}X 
			l c @{}}
		\toprule
		\textbf{Optimization Task} & \textbf{Retrieved Paper Title ($\mathcal{D}_{pdf}$)} & \textbf{Venue} & \textbf{Algorithm} \\
		\midrule
		High-Dimensional & Scalable Global Optimization via Local Bayesian Optimization & NeurIPS 2019 & TuRBO \\
		Large-Scale & Efficient Large-Scale Expensive Optimization via Surrogate-Assisted Sub-Problem Selection & TEVC 2025 & LSEO-S3 \\
		Real-World Low-Dimensional & Unexpected Improvements to Expected Improvement for Bayesian Optimization & NeurIPS 2023 & LogEI \\
		\bottomrule
	\end{tabularx}
	\caption{The retrieved SOTA-level literature ($\mathcal{D}_{pdf}$) across three diverse optimization tasks.}
	\label{tab:retrieved_papers}
\end{table*}

As comprehensively reported in the main text (see Section \ref{result}), we directly evaluated the native performance of these retrieved algorithms (i.e., TuRBO, LSEO-S3, and LogEI) as our representative baselines. The empirical results clearly indicate that while these SOTA-level solvers provide a strong algorithmic foundation, their overall performance is consistently inferior to the final customized solvers generated by AutoSG. This performance gap explicitly highlights the necessity and superiority of our proposed one-step self-refinement operator, which successfully introduces task-specific algorithmic improvements while strictly preserving the mathematical soundness of the original methods.

Furthermore, to ensure transparency and the full reproducibility of our empirical evaluations, the complete executable Python source codes of the optimal solvers generated by AutoSG across all three tasks ($\mathcal{S}^*_{\text{BBOB}, 1}$ , $\mathcal{S}^*_{\text{CEC}, 1}$ , and $\mathcal{S}^*_{\text{HPT}, 1}$ are made publicly available. These artifacts can be accessed via the following anonymous GitHub repository: \url{https://anonymous.4open.science/r/AutoSG-1CD4/}.

\end{document}